\PassOptionsToPackage{dvipsnames,table}{xcolor}
\documentclass[sigconf=true, nonacm=true, review=false, anonymous = false]{acmart}
\usepackage{soul}

\usepackage{booktabs} 
\usepackage{colortbl} 

\graphicspath{{pics/}}

\usepackage{etoolbox}
\usepackage{amsmath}
\usepackage{amsthm}
\usepackage{xspace}
\usepackage[table]{xcolor}
\usepackage{dsfont}
\usepackage{rotating}
\usepackage{bm}
\usepackage{subcaption}
\usepackage{hyperref}
\usepackage{mathtools}

\usepackage[algo2e,ruled,vlined,linesnumbered]{algorithm2e}
\usepackage{url}
\usepackage[noabbrev,nameinlink]{cleveref}

\crefname{prop}{property}{properties}
\creflabelformat{prop}{#2{\upshape(#1)}#3}

\crefname{pol}{policy}{policies}
\creflabelformat{pol}{#2{\upshape(#1)}#3}

\makeatletter
\def\NAT@spacechar{~}
\makeatother

\DeclareMathOperator{\Bin}{Bin}

\DeclarePairedDelimiter{\round}{\lfloor}{\rceil}

\newcommand{\oll}{$(1 + (\lambda,\lambda))$~GA\xspace}
\newcommand{\onell}{\oll}

\newcommand{\onemax}{\textsc{OneMax}\xspace}
\newcommand{\leadingones}{\textsc{LeadingOnes}\xspace}
\newcommand{\lbd}{$\lambda$\xspace}

\newcommand{\irace}{\textsc{irace}\xspace}

\newcommand{\theorydyn}{\textsc{theory}\xspace}
\newcommand{\binnedtheorydyn}{\textsc{binned\_theory}\xspace}

\newcommand{\tunedstatic}{\textsc{tuned\_static}\xspace}
\newcommand{\tuneddyn}{\textsc{tuned}\xspace}
\newcommand{\tuneddynbin}[1][]{\textsc{tuned\_bin#1}\xspace}
\newcommand{\tuneddynbincas}{\textsc{tuned\_cas\_bin}\xspace}

\DeclareMathOperator{\flip}{flip}
\DeclareMathOperator{\cross}{cross}

\newcommand{\assign}{\leftarrow}

\definecolor{cyan}{RGB}{0, 128, 128}

\allowdisplaybreaks

\begin{document}


\title{Using Automated Algorithm Configuration for Parameter Control}

\author{Deyao Chen}
\orcid{0009-0006-3210-9192}
\affiliation{
\institution{University of St Andrews}
\city{St Andrews}
\country{United Kingdom}
}

\author{Maxim Buzdalov}
\orcid{0000-0002-7120-8824}
\affiliation{
\institution{Aberystwyth University}
\city{Aberystwyth}
\country{United Kingdom}
}

\author{Carola Doerr}
\orcid{0000-0002-4981-3227}
\affiliation{
\institution{Sorbonne Universit{\'e}, CNRS, LIP6}
\city{Paris}
\country{France}
}

\author{Nguyen Dang}
\orcid{0000-0002-2693-6953}
\affiliation{
\institution{University of St Andrews}
\city{St Andrews}
\country{United Kingdom}
}

 \begin{abstract}
 Dynamic Algorithm Configuration (DAC) tackles the question of how to automatically learn policies to control parameters of algorithms in a data-driven fashion. This question has received considerable attention from the evolutionary community in recent years. Having a good benchmark collection to gain structural understanding on the effectiveness and limitations of different solution methods for DAC is therefore strongly desirable. Following recent work on proposing DAC benchmarks with well-understood theoretical properties and ground truth information, in this work, we suggest as a new DAC benchmark the controlling of the key parameter $\lambda$ in the $(1+(\lambda,\lambda))$~Genetic Algorithm for solving OneMax problems. We conduct a study on how to solve the DAC problem via the use of (static) automated algorithm configuration on the benchmark, and propose techniques to significantly improve the performance of the approach. Our approach is able to consistently outperform the default parameter control policy of the benchmark derived from previous theoretical work on sufficiently large problem sizes. We also present new findings on the landscape of the parameter-control search policies and propose methods to compute stronger baselines for the benchmark via numerical approximations of the true optimal policies.
 \end{abstract}


\maketitle

\section{Introduction}
\label{sec:intro}

Evolutionary algorithms and similar randomized search heuristics have a number of parameters that allow to fine-tune their behavior to the problem at hand and to the current stage of the optimization process. However, despite a long series of works studying which parameters work well for which algorithms and which problem scenarios, the question how to control these parameters in an automated fashion is wide open. Concepts such as \textit{hyper-heuristics}~\cite{BurkeGHKOOQ13}, \textit{parameter control}~\cite{EibenHM99,KarafotiasHE15,DoerrD18chapter}, and \textit{adaptive operator selection}~\cite{FialhoCSS10journal} dominate research on this topic in evolutionary computation (EC). 
More recently, approaches based on machine learning (ML), and in particular based on reinforcement learning (RL), propose to explicitly \emph{train} control policies. That is, where EC assumes that a given problem instance has to be solved instantly, the ML-approaches focus on settings in which similar problem instances need to be solved, and an explicit training is possible. This setting was first studied in~\cite{ManuelGECCO2019DDQN} and later named \emph{dynamic algorithm configuration} (DAC) in~\cite{BiedenkappBEHL20DACECAI}. It quickly became a very active area of research~\cite{shala-ppsn20,BiedenkappHeuristicSelection,Adriansen2022DACjournal,BiedenkappDKHD22GECCO,DACBench,xue2022multi,tessari2022reinforcement}. A parallel research direction in the hyper-heuristic community is on \textit{automated algorithm design,} which focuses on leveraging deep RL approaches for training policies to select the best algorithm components based on the search states~\cite{zhang2022deep,yi2022automated,yi2023automated}. 

Regardless of whether adopting the EC view or whether considering the DAC setting, it is highly desirable to have access to benchmarks that support a sound investigation of the proposed methods. As in other branches of computer science, benchmarking helps to compare the efficiency of different algorithms and to investigate their strengths and weaknesses. Benchmarking has a long tradition in EC, and has undoubtedly helped our community mature~\cite{hansen2020cocoJournal,TBB20benchmarking}. In difference to \textit{competitive testing}, where the main focus is on the performance of the solvers, \textit{benchmarking} relies on settings that are sufficiently well understood to help us gain insight into the working principles that drive this performance -- knowledge that not only supports the improvement of the algorithms but also leads to a better deployment of the existing approaches in practice.

The idea of leveraging settings with well-understood theoretical properties and using them for benchmarking DAC approaches was recently promoted in~\cite{BiedenkappDKHD22GECCO}. The paper adopted the (1+1) Randomized Local Search for solving variants of the \leadingones problem with configurable problem dimension and action spaces, where the action space corresponds to the portfolio of the parameters that the DAC policies are allowed to choose from. Notably, the paper~\cite{BiedenkappDKHD22GECCO} extended previously known theoretical results from~\cite{DoerrW18,Doerr19domi} to settings with restricted portfolio. 
Using these provably optimal \textit{``ground truth''} policies, their work demonstrated that an off-the-shelf RL-based learning mechanism can work very well if both the number of parameter values to choose from and the problem dimension are rather small. As soon as either of them increases beyond some relatively moderate threshold, the observed performance deteriorated quickly.  

\textbf{Our contributions.} Following the approach in~\cite{BiedenkappDKHD22GECCO}, in this paper we propose a new DAC benchmark derived from EC theory. An essential property of a DAC benchmark is the clear discrepancy between static configuration and dynamic control policies. To that end, we adopt the setting proposed in~\cite{DoerrDE15}, which involves controlling the population size parameter $\lambda$ in the $(1+(\lambda,\lambda))$~Genetic Algorithm (GA) for the \onemax problem. As in the \leadingones benchmarks, the goal is to minimize the expected running time of the algorithm, i.e., the total number of evaluations that the algorithm performs before it reaches an optimal solution. The setting also comes with an asymptotically optimal parameter control policy with linear expected runtime~\cite{DoerrDE15}, which will be used as a baseline for our DAC experiments in this paper.

The new benchmark has a number of interesting properties compared to the \leadingones. First, the performance gain obtained by an optimally controlled parameter setting over the best possible static setting is $\omega(1)$, i.e., it grows more than a constant factor as the problem dimension increases, while the advantage of an optimally controlled policy for \leadingones is ``only'' a constant factor, i.e., $\Theta(1)$. Second, as we will show in our analysis~(\Cref{sec:exact:landscape}), the parameter search space has an unusual landscape with frequent abrupt changes across the whole area, making it potentially challenging for DAC approaches. 

It is worth to note that deep RL is not the only solution method for DAC. An alternative approach is to formulate DAC as a (meta-)optimization problem, where we search in the policy space directly via black-box optimization methods~\cite{Adriansen2022DACjournal}. Our second contribution is on this direction: we investigate the potential of using automated algorithm configuration techniques to solve DAC problems. More specifically, we formulate a policy of our benchmark as a static parameter setting and configure it using the automated algorithm configuration tool \irace~\cite{irace}. As will be detailed in~\Cref{sec:irace}, a naive application of \irace does not work well, especially when the problem size is over $200$, which is considered rather small for \onemax problems. This can be accounted to the large number of parameters the tuning has to deal with and the complicated landscape of the parameter search space. We then propose a \emph{binning} approach to reduce the search space, and a \emph{cascading} application of the tuning. A combination of those two strategies results in significant improvement in performance. The tuning is able to find policies that consistently outperform the theory-derived policy in~\cite{DoerrDE15}.

Our last contribution involves expanding our understandings on the proposed DAC benchmark. We extend recent works on numerically computing optimal control policies for concrete dimensions via dynamic programming. Originally proposed in~\cite{BuskulicD21} and later refined in~\cite{BuzdalovD20,AntonovBBD21}, this approach allows us to approximate the optimal control policy of our new benchmark with very good precision. In this work, we compute the approximations of both the generally optimal policies and the optimal binning policies, which show that there is a consistent but small gap between the two. Apart from serving us as new baselines for our tuning-based approach and for other DAC methods in general, our work also reveals some facts about the \oll that we expect to be of interest in the context of running time analysis, parameter setting, and algorithm design. Indeed, search landscapes associated with parameter setting problems are usually thought to be rather smooth~\cite{PushakH18PPSN,PushakH22telo}, but our findings suggest that it is not always true, and in particular not for parameter control settings. This finding reinforces a similar observation made in~\cite{AntonovBBD21} for landscape of parameter control policies for the much simpler $(1+\lambda)$-type evolutionary algorithms. We consider such insights on (un)structured parameter control landscapes very valuable for further algorithm design -- whether in the context of DAC or in the classic black-box optimization setting. 

\textbf{Availability of code and data.} To adhere to reproducibility standards discussed in~\cite{Reproducibility21TELO}, our code and data are available at~\cite{data}.

\section{Controlling the Population size of the \texorpdfstring{$(1+(\lambda,\lambda))$~GA}{1LL-GA} on OneMax}
\label{sec:background}

\textbf{Notation.} We always denote by $n$ the dimension of the search space. For a search point $x\in \{0,1\}^n$ we write $x=(x_1, \ldots, x_n)$. For two real number $a$ and $b$ we denote by $[a..b]$ the set of all integers $k$ that satisfy $a\le k \le b$. 

\textbf{OneMax.} 
In this work, we are interested in minimizing the expected optimization time of the \oll on \onemax. \onemax is one of the most important benchmark problems in the analysis of evolutionary algorithms (EAs). It denotes the collection of functions 
$\{f_z \mid z \in \{0,1\}^n\}$ with 
$f_z:\{0,1\}^n \to [0..n], x \mapsto |\{i \in [1..n] \mid x_i=z_i\}|$, the function that counts in how many positions the solution candidate $x$ agrees with the secret target string $z$. As it was discussed in~\cite{DoerrW14memory}, \onemax can be seen as the Mastermind problem with two colors, 0 and 1. The name ``\onemax'' originates in the fact that for analyzing the performance of so-called \textit{unbiased} algorithms, it suffices to study their behavior on the function $f_{(1, \ldots, 1)}$, see~\cite{LehreW12,Doerr20chapter} for detailed explanations. This is also the case for the \oll introduced below.  
Despite  its simplicity, the \onemax problem is highly relevant for understanding the behavior of EAs in regimes in which we have a good fitness-distance correlation. That is, \onemax helps us understand how algorithms perform in environments in which they are not misled by search points that have better fitness value than another while at the same time being farther away from the optimum. 

\textbf{Background on the \oll.} 
The \oll was originally introduced in~\cite{DoerrDE15} to formally prove that the use of crossover can be beneficial even for optimizing rather simple functions such as \onemax. It later started to play an important role in the analysis of parameter control mechanisms.  Already in~\cite{DoerrDE15} it was shown that a fitness-dependent setting of the key parameter $\lambda$ can lead to expected running times that are asymptotically better than that of the best static setting. Later works showed that a simple one-fifth success rule, as adopted to the discrete setting in~\cite{KernMHBOK04}, leads to best possible linear running time. This result inspired a number of follow-up works on parameter control~\cite{DoerrD18chapter}. 
More recently, the \oll plays another important role for the study of unusual mutation operators such as the heavy-tailed ``fast'' mutation operators introduced in~\cite{fastGA}; see~\cite{AntipovBD22} for a recent example. Finally, the \oll has also been studied in other contexts, see~\cite{AntipovDK19LO} for work on \leadingones, \cite{BuzdalovD17,GoldmanP15} for applications to satisfiability problems, and \cite{AntipovDK22} for more results and further references. 

\textbf{The \oll.} 
Alg.~\ref{alg:oll} presents the pseudo-code of the \oll version analyzed in this work. The algorithm is initialized by sampling and evaluating a search point chosen from $\{0,1\}^n$ uniformly at random (u.a.r.). It then proceeds in rounds, consisting of one mutation phase and one crossover phase each. In the mutation phase, $\Lambda = \round{\lambda}$ new solution candidates (``\emph{offspring}'') are sampled, where $\round{\lambda}:=\lfloor \lambda \rfloor$ if $\lambda-\lfloor \lambda \rfloor < 0.5$  and $\round{\lambda}:=\lceil \lambda \rceil$ otherwise. Each offspring is sampled by inverting the bits in $\ell$ position that are chosen u.a.r.. Here, $\ell$ is a random number that is kept fixed throughout one iteration and that is sampled from the \emph{resampling} binomial distribution $\Bin_{>0}(n,\lambda/n)$ that performs $n$ trials with success rate $\lambda/n$ each and resamples i.i.d. until a value greater than 0 is found. The best of these $\Lambda$ offspring (ties broken u.a.r.), referred to as $x'$ in Alg.~\ref{alg:oll}, is selected to participate in the crossover phase. In the crossover phase, another $\Lambda$ search points are generated, each sampled from a uniform crossover between the original parent $x$ and the selected mutant $x'$. The crossover operator $\cross_{1/\lambda}(x,x')$ treats each position independently and sets $y_i=x'_i$ with probability $1/\lambda$ and it sets $y_i=x_i$ otherwise. The so-created \emph{``crossover offspring''} are only evaluated if they are different from both parents, and they are not further taken into consideration otherwise. For $y'$ denoting the best of the samples generated in the crossover phase (ties broken again u.a.r.), we then let $y=y'$ if $y'$ is strictly better than $x'$ and we let $y=x'$ otherwise. Finally, $y$ replaces $x$ as parent for the next iteration if it is at least as good, i.e., if $f(y) \ge f(x)$.

\begin{algorithm2e}[t]%
\SetKwInOut{Input}{Input}
\Input{problem size $n$ \newline a parameter control policy $\pi: [0..n-1] \to [1;n]$}
\textbf{Initialization:} 
		Sample $x \in \{0,1\}^n$ u.a.r. and evaluate $f(x)$\;
\textbf{Optimization:}
\For{$t=1,2,3,\ldots$}{
Let $\lambda =\pi(f(x))$\;
\underline{\textbf{Mutation phase:}}\\
	Sample $\ell$ from $\Bin_{>0}(n,\lambda/n)$\;
    Set population size $\Lambda$ as $\round{\lambda}$\;
	\lFor{$i=1, \ldots, \Lambda$}
         {$x^{(i)} \assign \flip_{\ell}(x)$; Evaluate $f(x^{(i)})$}
	Choose $x' \in \{x^{(1)}, \ldots, x^{(\Lambda)}\}$ with $f(x')=\max\{f(x^{(1)}), \ldots, f(x^{(\Lambda)})\}$ u.a.r.\;
\underline{\textbf{Crossover phase:}}\\
\For{$i=1, \ldots, \Lambda$}
{$y^{(i)} \assign \cross_{1/\lambda}(x,x')$\; 
\lIf{$y^{(i)} \notin \{x,x'\}$}{evaluate $f(y^{(i)})$}}
Choose $y' \in \{y^{(1)}, \ldots, y^{(\Lambda)}\}$ with 
    $f(y') = \max\{f(y^{(1)}), \ldots, f(y^{(\Lambda)})\}$ u.a.r.\;
\underline{\textbf{Selection and update step:}}\\
\lIf{$f(y') > f(x')$}{$y \assign y'$ \textbf{ else } $y \assign x'$}
\lIf{$f(y)\ge f(x)$}{$x \assign y$} 
}
\caption{The \oll variant analyzed in this paper.
}
\label{alg:oll}
\end{algorithm2e}

\textbf{Parameter settings.} In the \textit{static} version of the \oll, the value of $\lambda$ is fixed throughout the whole run, i.e., the control policy $\pi$ in Alg.~\ref{alg:oll} is constant. In the \textit{dynamic} version, $\lambda$ can take different values at different stages of the optimization process. 

The default dynamic control policy that we compare our results against, namely \theorydyn, assigns to each fitness value $i$ the parameter value $\pi(i):= \sqrt{n/(n-i)}$. It was used in~\cite[Theorem~8]{DoerrDE15} to prove an asymptotic super-constant speedup over the best static choice of $\lambda$. 

\section{Configuring Control Policies }
\label{sec:irace}

\begin{figure}[h]
  \centering
  \includegraphics[width=0.5\textwidth]{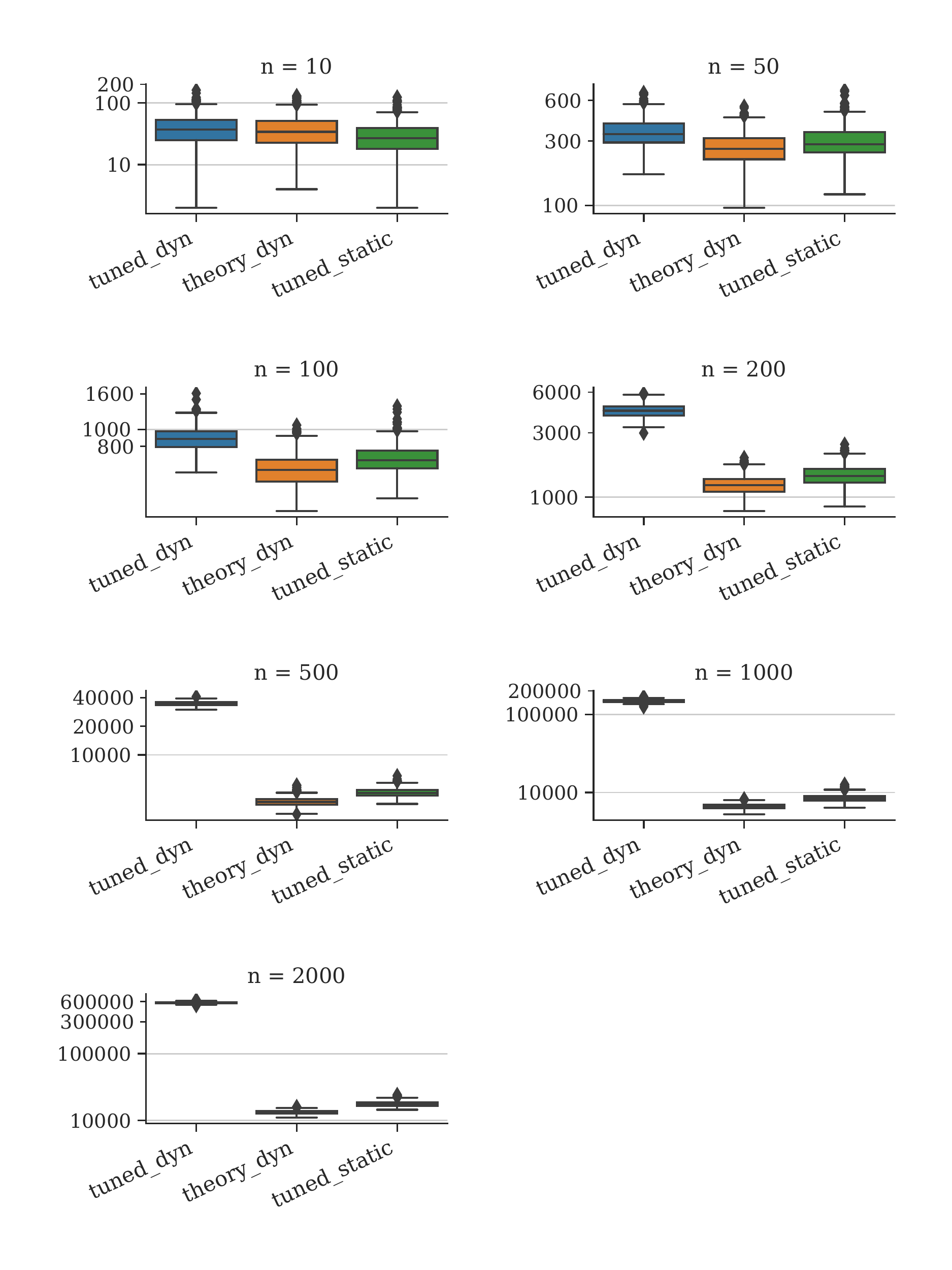}
  \caption{Performance of a naive dynamic tuning with \irace (\tuneddyn), a static tuning (\tunedstatic), and the default policy derived from theory (\theorydyn), evaluated across 500 random seeds.}
  \label{fig:dyn_vs_static}
\end{figure}

We represent a parameter control policy for the \oll as a mapping from a current fitness to a specific \lbd value. The problem of finding the best parameter control policy therefore can be stated as a static automated algorithm configuration, where the number of parameters is equal to the problem size $n$: each parameter correspond to the \lbd value of a fitness value in the range of $[0..n-1]$. We use the automated algorithm configuration tool \irace~\cite{irace} to solve this task. 
We apply this dynamic tuning approach, namely \tuneddyn, on various problem sizes from $10$ to $2000$.
The tuning budget is set as $50000$ runs. 
Since the performance metric is runtime related, the adaptive capping feature of \irace is enabled as it has been shown to significantly improve the tuning performance in various cases~\cite{caceres2017experimental,SouzaRL22iraceCapping}. 
As we are interested in optimizing \emph{expected} running time, we set \irace's statistical test to be the Student t-test. Moreover, to account for the noisy nature of the benchmark, the \emph{firstTest} parameter (the number of instances/seeds being evaluated at the beginning of each \irace's iteration before the first statistical test is applied) is increased from $5$ (default) to $10$. All other parameters of \irace are set as default. For comparison, we also apply \irace on a \emph{static} version of the \onell where \lbd is fixed during the whole run. 
This version is named \tunedstatic, and the tuning budget is set as
min($100 \times n$, 20000) runs. 
The final configuration found by each tuning experiment is evaluated across $500$ different random seeds, and their performances are presented in~\Cref{fig:dyn_vs_static}.

\begin{figure*}[!t]
  \centering
  \includegraphics[width=\textwidth]{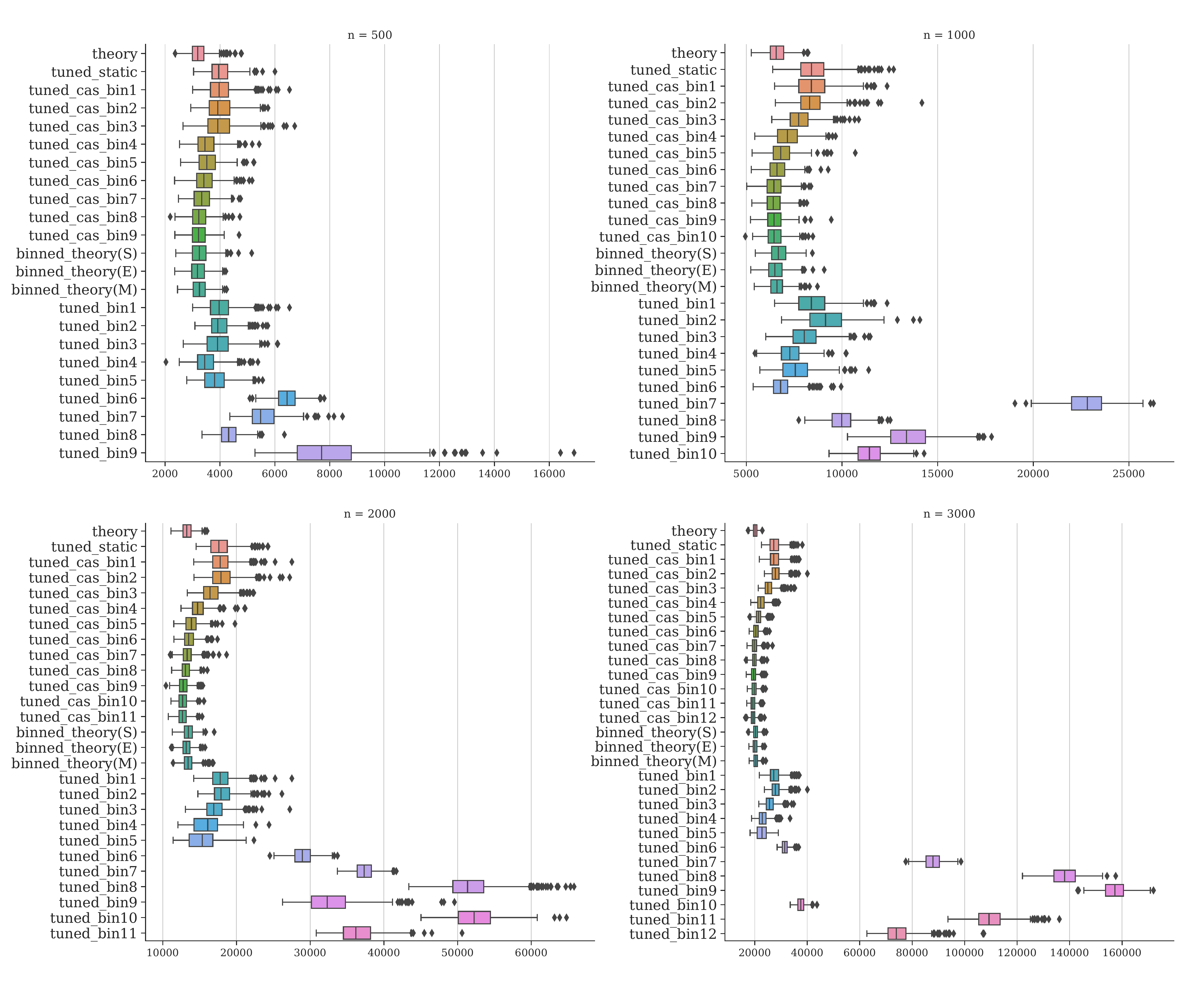}
  \caption{Performance (in terms of function evaluations) of dynamic tuning with binning (\tuneddynbin), and with binning and cascading (\tuneddynbincas). The number at the end of each configuration represents the number of bins. Other baselines include: \theorydyn and its binned versions, and the static tuning (\tunedstatic). All configurations are evaluated across 500 random seeds.}
  \label{fig:all_tuned}
\end{figure*}

It is clear that the dynamic tuning approach does not scale well with problem sizes: the expected runtime of \tuneddyn is much higher 
than both \theorydyn and \tunedstatic, especially when $n \geq 200$.~\footnote{for $n=10$, the average performance of \tunedstatic is better than both \theorydyn and \tuneddyn. But this is likely due to the problem size being too small to see the impact of dynamic parameter control.} The differences in performance (i.e., \tuneddyn vs \theorydyn and \tuneddyn vs \tunedstatic) are statistically significant~\footnote{In this paper, whenever multiple statistical tests are conducted, the Bonferroni correction is applied.} for $n \geq 50$ according to the Wilcoxon rank sum test with a confidence level of $99.9\%$.   
One possible explanation for the bad performance of \tuneddyn is due to the large number of numerical parameters \irace has to deal with. In fact, automated algorithm configuration scenarios often either: (i) involves a few dozens parameters at most, or in some cases, up to $200$ parameters; or (ii) has a large number of categorical parameters with only a few possible values~\cite{hutter2014aclib}. 

\begin{figure}[!t]
  \centering
  \includegraphics[width=0.4\textwidth]{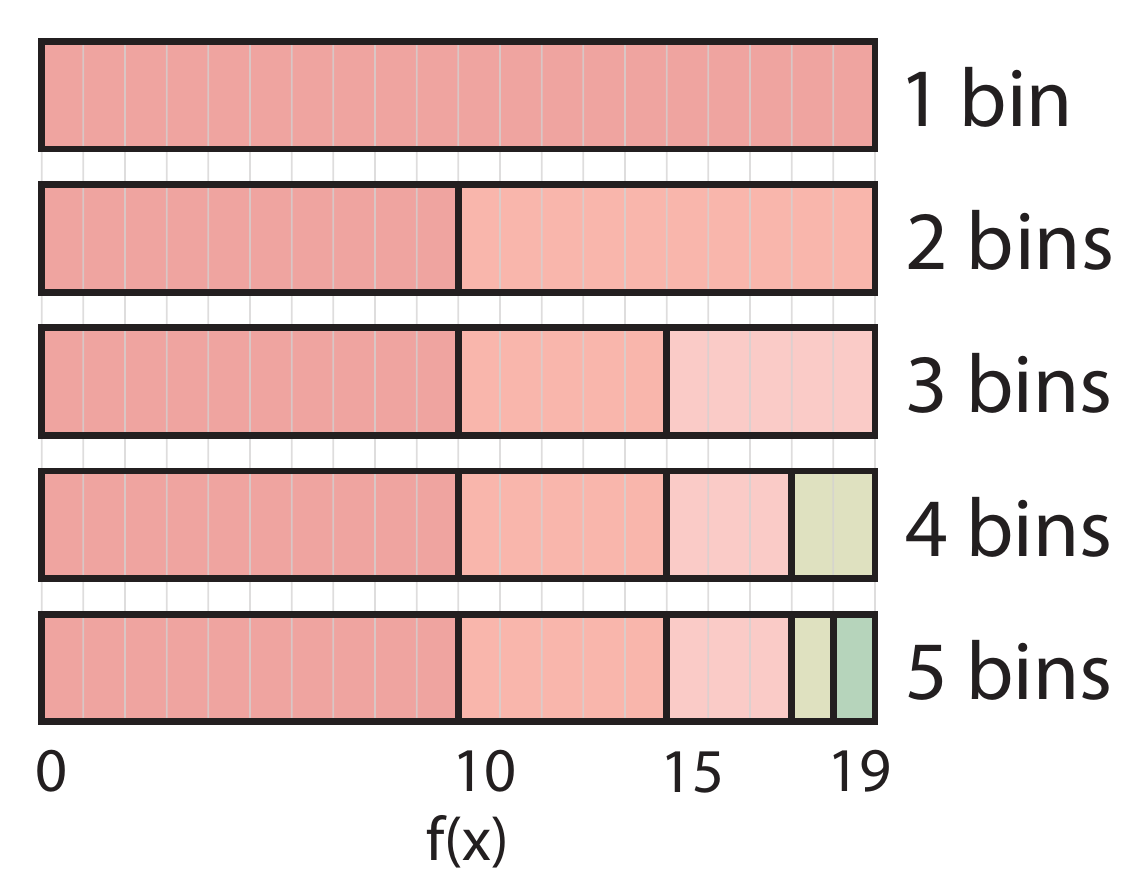}
  \caption{An illustration of the division of the bins for $n = 20$.}
  \label{fig:bin_illustration}
\end{figure}

To reduce the parameter space, we define a \emph{binning} approach where we partition the objective space $[0..n]$ into $k$ consecutive bins \{$B_1$, $B_2$, .., $B_k$\}, and only tune one \lbd parameter for all fitness values in the same bin, i.e., the number of parameters is now reduced from $n$ to $k$. In many optimization problems, the closer we get to the optimal, the harder it is to optimize. Therefore, we define the bins so that their sizes are gradually reduced as the objective values increased. More concretely, we set $B_i=[a_i .. a_{i+1}-1]$ ($i \in [1,k-1]$) where $a_i = n - \lfloor \frac{n}{2^{i-1}} \rfloor$ $\forall i \in [1,k]$, and $B_k=[a_k..n]$.~\Cref{fig:bin_illustration} illustrates how the bins are created with different $k$ values. 
We name this approach \tuneddynbin. From this point onward, we will focus on sufficiently large problem sizes ($n \geq 500$) as those are the cases where the discrepancy between the static and the dynamic policies is obvious. The tuning budget is set as $5000$, $10000$, $20000$, and $21000$ for $n$ equals to $500$, $1000$, $2000$, and $3000$, respectively. 

A larger number of bins results in more flexibility in the tuning and better quality of the best possible parameter-control policy, but it may also lead to more difficulty in searching in such space due to its size. To study the effect of this choice, we conduct experiments with all possible numbers of bins ($k \in [1..\lceil \log_2{n} \rceil]$). The performance of the final configurations are presented in~\Cref{fig:all_tuned}. We also conducted another baseline for the comparison: the binned versions of the \theorydyn, namely \binnedtheorydyn, where the fitness range is split into $\lceil \log_2{n} \rceil$ bins, and all \lbd values belonging to the same bin are derived from \theorydyn at the start(S), the middle (M), or the end(E) of the bin (for the middle case, if the number of elements in a bin is even, we choose the smaller point). 

Results in~\Cref{fig:all_tuned} clearly indicates a trade-off between the number of bins and the tuning performance. Starting from one bin, the tuning performance generally improves when the number of bins is increased until it reaches a certain limit (around $5$ or $6$). The tuning results after such point not only drastically degrades, but are also \emph{unstable}. Consider, for example, the case of $n=500$, we have both \tuneddynbin{6} and \tuneddynbin{7} performing quite badly compared to the cases with smaller numbers of bins, indicating degradation of the tuning performance, and we would expect \tuneddynbin{7} to be worse than \tuneddynbin{6} due to the larger number of parameters \irace has to deal with. However, we observe the opposite. 
To check whether the counter-intuitive performance is due to the instability of the tuning, we repeat both tuning experiments $5$ times. And in fact, the average runtime of \tuneddynbin{6} reduces from $6446$ to $4921$, while the runtime of \tuneddynbin{7} increases from $5589$ to $6043$.

To improve the tuning further, we propose a \emph{cascading} strategy, where the final configuration of the tuning with $k$ bins is given as an initial configuration for the tuning with $k+1$ bins. The cascading gives each tuning a head start by leveraging results of the previous step. The new tuning experiments are named \tuneddynbincas, and their results are shown in~\Cref{fig:all_tuned}. The tuning performance is improved drastically. And the final configuration (found with the largest number of bins) for each problem size even statistically significantly outperforms \theorydyn (both original and binned versions) for $n \geq 1000$ according to the Wilcoxon rank sum test with a confidence level of $99.9\%$. Note that for $n=500$, the two configurations are not statistically significantly different. 

One may argue that the good performance of the cascading approach may come from the fact that the total tuning budget is larger than each individual \tuneddynbin experiment, since with cascading, the tunings are executed sequentially until we reach the largest number of bins. To investigate this point further, we re-run the last \tuneddynbin experiments for both $n \in {500, 1000}$ with $10$ times of their original budget (which is roughly the same as the total budget for all \tuneddynbincas experiments of the same problem size). We observe some improvement in performance for $n=500$: average runtime of the final configuration reduces from $7988$ to $5818$, but there is still a large gap compared to the \tuneddynbincas with an average runtime of $3234$. Interestingly, for $n=1000$, the new configuration is even worse than before (original runtime: $11468$, new runtime: $15109$, \tuneddynbincas: $6492$). Again, this can be accounted to the instability of dynamic tuning without cascading. Those observation confirms the clear advantage that cascading offers to the tuning performance.

The results of this section indicates the potential of applying automated algorithm configuration for the dynamic tuning of the \onell on \onemax. Our binning and cascading strategies help to find configurations that are better than the best known control policy derived from theory. This is particularly encouraging, given that the tuning does not have access to \theorydyn. The next question would be: has the tuning been able to find the best possible policies, or is there still room for improvement? To answer such question, in the next section, we will describe a method to compute stronger baselines for this benchmark.

\section{Exact Computations}

In this section we describe an algorithm of computing the exact expected running time
of the \onell on \onemax provided we are given the mapping from the parent fitness $f$ to the parameter value $\lambda = \lambda(f)$. Based on that, we perform some basic landscape analysis and propose a way to closely approximate
the best baseline policies (the truly best policy and the best one among the binned policies) using numeric minimization.

\subsection{Computing Runtime for Given Parameters}\label{sec:exact:dp}

Let the current parent's \onemax fitness be $f$, and the current parameter be $\lambda$.
Let $\Lambda$ be the derived population size; the actual algorithm uses $\Lambda = \round{\lambda}$,
but we are going to use different values during landscape analysis. As the \onell is elitist and all parents with the same fitness produce stochastically identical behavior, we can use the dynamic programming approach similar to the one used in previous works~\cite{BuskulicD21,BuzdalovD20,BuzdalovD21,AntonovBBD21}. Basically, we compute the remaining runtimes backwards from the optimum: assuming that $T_f$ is the expected time to reach the optimum when the parent's fitness is $f$, we note that $T_n = 0$ and for every $f$ the value of $T_f$ depends only on $T_{f+1}, T_{f+2}$ and other values for higher fitness.

To process the fitness value $f$, our algorithm essentially considers all possible event chains that the \onell performs, computes their probabilities and saves the computational effort when possible. Since most of the probabilities are fractions formed by binomial coefficients, each of which may be small or large, we use the standard trick of computing logarithms of those quantities (which are aggressively cached to avoid calling the expensive logarithm function) and taking an exponent of them only when summing.

We start with considering how many bits are flipped in the mutation phase. The probability of flipping exactly $\ell$ bits follows from the fact that $\ell \sim \Bin_{>0}(n, \lambda / n)$. Let $g$ be the number of ``good'' bits in the best offspring, that is, how many out of the $\ell$ bits are flipped from 0 to 1 in $x'$. The probability $P[g \mid \ell]$
can be computed as follows: let $P_m(i)$ be the (easily computed) probability to have $i$ ``good'' bits in one offspring, then
\begin{equation*}
    P[g \mid \ell] = \left(\sum\nolimits_{i=0}^{g} P_m(i)\right)^{\Lambda} 
                   - \left(\sum\nolimits_{i=0}^{g-1} P_m(i)\right)^{\Lambda}
\end{equation*}
by combinatorial arguments. Hence, the crossover phase starts with $x'$ being different from the parent $x$ in $\ell$ bits, $g$ of which are ``good''.

Each crossover offspring is obtained from the parent $x$ by considering the $\ell$ bits that are different between $x$ and $x'$ and flipping $\delta_g$ of those bits from 0 to 1, and $\delta_b$ bits from 1 to 0. Again, remembering that only $g$ of these $\ell$ bits have the value 1 in $x'$, it holds that $0 \le \delta_g \le g$ and $0 \le \delta_b \le \ell - g$.
The probability $P[\delta_g, \delta_b | \ell, g]$ can again be computed by combinatorial arguments. The fitness of such an offspring is $f + \delta_g - \delta_b$. Since the algorithm's state changes only when the fitness improves, and because the best mutation offspring, $x'$, can also replace the parent, we are rather interested in the truncated fitness change, e.g. $\delta = \max\{0, f(x') - f, \delta_g - \delta_b\}$. Let $P_c(i)$ be the probability of obtaining such truncated fitness change of $i$ in one crossover offspring, then
\begin{equation*}
    P[\delta \mid \ell, g] = \left(\sum\nolimits_{i=0}^{\delta} P_c(i)\right)^{\Lambda}
                           - \left(\sum\nolimits_{i=0}^{\delta-1} P_c(i)\right)^{\Lambda}
\end{equation*}
similarly to the above. If $\delta = 0$, we consider the iteration a failure. Otherwise the algorithm updates the fitness of the parent, for which we already computed the runtime $T_{f + \delta}$. We sum up these remaining expected times, weighed by the probabilities of all the preceding events to happen, as well as these probabilities themselves.

Getting back to the current fitness value $f$, we obtain the probability of improving the parent $p_i$ and the conditioned remaining expected time $t_i$. Denote the expected iteration cost, in fitness evaluations, to be $\tau_{\lambda}$. Then, $T_f = \tau_{\lambda} + p_i \cdot t_i + (1 - p_i) T_f$, which resolves to a well-known solution $T_f = (\tau_{\lambda} + t_i) / p_i$. The value of $\tau_{\lambda}$ is a sum of $\Lambda$ evaluations in the mutation phase and $\Lambda \cdot (1 - (1 / \lambda)^{\ell} - (1 - 1 / \lambda)^{\ell}$) evaluations in the crossover phase that do not produce a crossover offspring equal to either of its parents. The final runtime of the policy is obtained from all $T_f$, given that the algorithm is initialized with fitness $f$ with the probability $\binom{n}{f} / 2^n$.

The overall runtime of the whole process can be estimated as $O(n^4)$ assuming a straightforward implementation. However, we perform certain optimizations in order to be able to work with problem sizes as large as $n=2000$ in reasonable time.

\textbf{Implementation detail 1}. Note that the probabilities $P[\delta \mid \ell, g]$ do not depend on the current fitness, but do depend on both $\lambda$ and $\Lambda$. For this reason, they can be computed once and stored for later re-use. Since the probabilities for a tuple $(\ell, g, \lambda, \Lambda)$ are computed in $\Theta(g^2)$ time and take only $\Theta(g)$ to store and use, such caching may improve the overall computation time by a factor up to $O(n)$. For certain scenarios the overall size of cached probabilities exceeds the amount of memory available, so we use the adaptive cache that retains only the most expensive entries w.r.t.~computation time.

\textbf{Implementation detail 2}. To improve the computation time further, we ignore the events as long as their probability does not change the existing sum of probabilities in the machine precision when added to it, that is, whenever $p + \delta p = p$ in the 64-bit floating point type commonly known as \texttt{double}. Where possible, we reorganize computations so that the major contributions are evaluated first to further accelerate computation.

\subsection{Landscape Analysis}\label{sec:exact:landscape}

\begin{figure}[!t]
\includegraphics{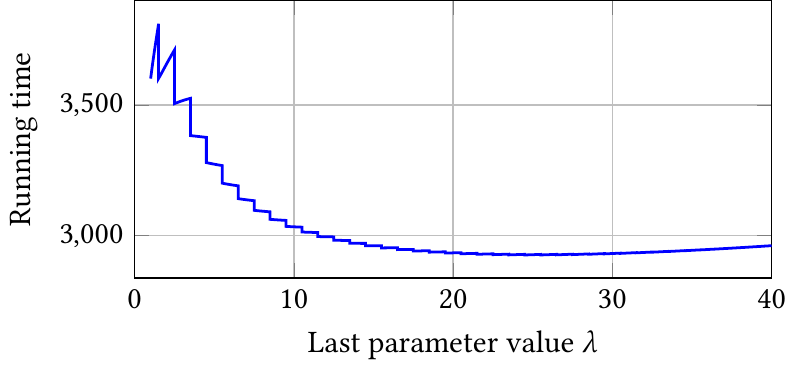}\par
\includegraphics{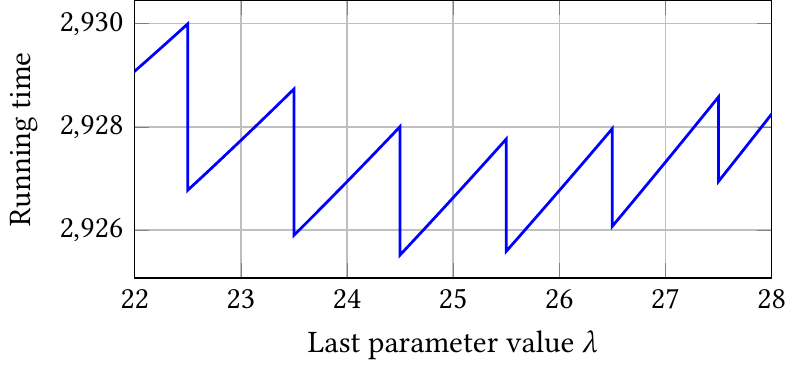}
\caption{Example runtimes for $n=500$, fixed binned policy and different values of the last $\lambda$}
\Description{Example runtimes for n=500, fixed binned policy and different values of the last lambda}
\label{fig:landscape:1}
\end{figure}

The algorithm outlined in the previous section makes it possible to conduct a simplified form of the landscape analysis.

Our first setting uses $n=500$ and employs binning to assign parameter values to fitness values. For all but the last bin we use the following sequence of values: [1.0, 1.0, 1.0, 1.0, 6.5, 8.5, 11.5, 16.5], which, as we show later, is the best known binned policy for this problem size. For the last bin, which corresponds to the fitness value of $n-1=499$, we test all possible values in the range $[1; 40]$ with a step of $0.1$. Additionally, for every half-integer value (that is, $x + 0.5$ for an integer $x$, which rounds to $x+1$) we also use the largest value that is smaller than it and is representable in the machine precision (such a value rounds to $x$).

The plot of the resulting expected runtimes, computed as detailed in the previous section, is shown in~\Cref{fig:landscape:1}. One can easily observe the saw-like shape of the plot, which introduces multiple local optima and a complicated search space even when looking for one parameter. Note that the abrupt changes happen at the half-integer values, which correspond to the points when the population size, which is $\round{\lambda}$, changes as the parameter $\lambda$ increases by a negligible amount. Note that a similar thing happens for every other bin value, and even if an arbitrary-shaped policy is sought for.

\begin{figure}[!t]
\includegraphics[width=0.4\textwidth]{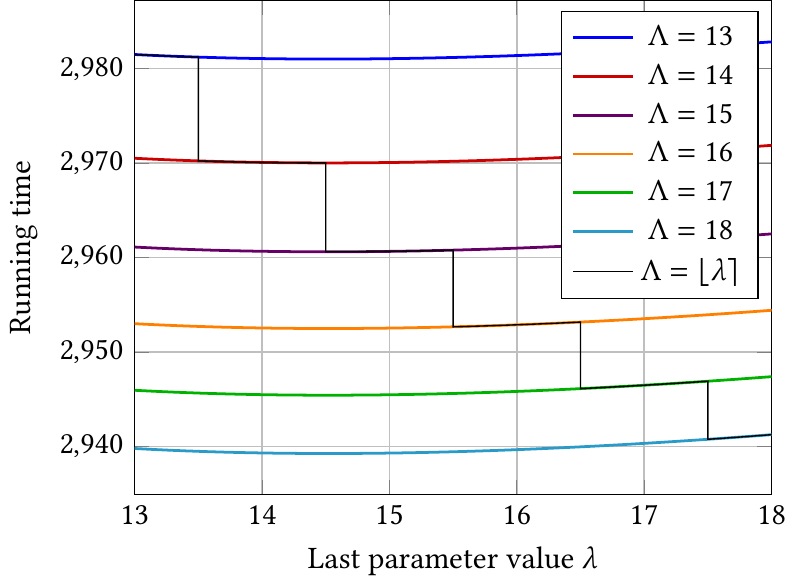}
\caption{Example runtimes for $n=500$, fixed binned policy, and different values of the last $\lambda$ and population size $\Lambda$}
\Description{Example runtimes for n=500, fixed binned policy and different values of the last lambda and population size}
\label{fig:landscape:2}
\end{figure}

To investigate this effect further, we consider the similar setting, but this time we vary both $\lambda$ and the population size $\Lambda$ independently. \Cref{fig:landscape:2} shows a closeup to the region of $\lambda\in[13;18]$ and population sizes $\Lambda\in[13..18]$. One can see that when the population size $\Lambda$ is fixed, the plot is smooth with respect to $\lambda$ as expected. However, when only the combinations $\Lambda = \round{\lambda}$ are considered, it can happen that the best $\lambda$ is at either of the interval endpoints or somewhere in the middle (all these cases are covered in~\Cref{fig:landscape:2}).

All in all, this effect makes it complicated to compute optimal policies, either numerically or analytically. What is more, it may contribute significantly to the difficulty for the parameter tuning problem observed in~\Cref{sec:irace} for \irace.

\begin{figure}[!t]
\includegraphics{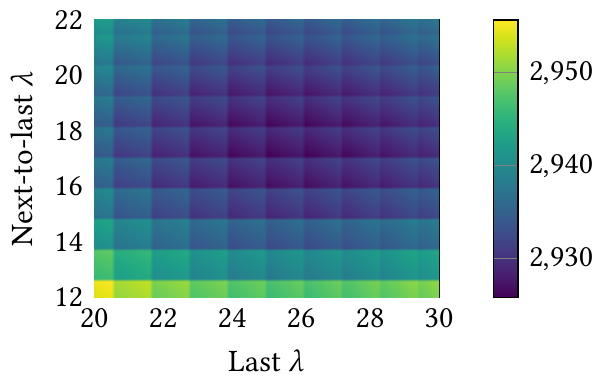}
\caption{Example runtimes for $n=500$, fixed binned policy, and different values of last two $\lambda$ values}
\Description{Example runtimes for n=500, fixed binned policy and different values of the last lambda values}
\label{fig:landscape:3}
\end{figure}

Our final setting investigates the effect of interaction between parameter values. Different to the first one, we consider changing the parameter values for the two last bins. The result is represented as a heatmap in~\Cref{fig:landscape:3} for parameters close to the optimal ones, which features a clear grid pattern. While within each cell one can observe gradual color change in directions that are not parallel to coordinate axes, indicating nonlinear interactions between the parameters that might be hard to capture, it is also clear that the effects from rounding~--- that is, multidimensional ruggedness of the landscape~--- are at least as large.

\subsection{Computing the Best Policies}\label{sec:exact:policies}

In order to compute the best mapping $\lambda(f)$ from fitness values to parameters values, given the problem size $n$, we modify the dynamic programming approach detailed in~\Cref{sec:exact:dp} to compute the best $\lambda$ for a given fitness $f$ assuming this has been done for all higher fitness values. This is done in line with the previous research~\cite{BuskulicD21,BuzdalovD20}, however, due to the nature of the dependency on parameters found in~\Cref{sec:exact:landscape}, a more complicated optimization problem needs to be solved for each fitness value.

Based on the preliminary computations involving smaller problem sizes, we consider intervals of the form $[x - 0.5; x + 0.5)$ for integer $x$, with an exception of the first and last intervals $[1; 1.5)$ and $[n - 0.5; n]$. The intervals open from the above are treated as closed intervals with the upper boundary replaced by the largest 64-bit floating-point value strictly smaller than that boundary. Within each interval $[\lambda_1, \lambda_2]$, we evaluate four values $\lambda_1$, $\lambda_1 + \varepsilon$, $\lambda_2 - \varepsilon$, $\lambda_2$ with $\varepsilon = 10^{-8}$ to determine whether the optimum is on the boundary or within the interval. In the latter case, we run a variant of ternary search suitable for parallel evaluation to find the best value numerically. Finally, to save computational efforts further, we reduce the number of intervals checked by observing that the optimum is in one of the intervals corresponding to:
\begin{itemize}
    \item population size 1 or 2;
    \item the end of a sequence of population sizes, starting from 2 upwards, such that the runtimes decrease, which is tested if the runtime for population size 3 is smaller than for population size 2 (this case is typical for large fitness values);
    \item the end of a sequence of population sizes, starting from $n$ downwards, such that the runtimes decrease, which is tested if the runtimes for population sizes $n, n - n/8, n - n/4, n - n/2$ are not monotonically decreasing (this case is typical for small fitness values less than $n/3$).
\end{itemize}

\begin{figure}[!t]
\includegraphics[width=0.4\textwidth]{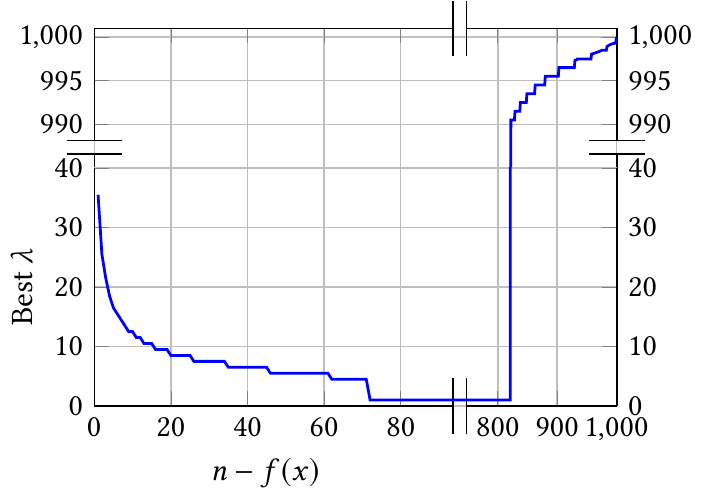}
\caption{Best dynamic policy for $n=1000$}
\Description{Best dynamic policy for n=1000}
\label{fig:best-dynamic}
\end{figure}

\Cref{fig:best-dynamic} shows the best dynamic policy for $n=1000$. There are clearly two regions with non-trivial best parameter values. The one closer to the optimum (which is on the left) is quite close to the theoretic value, $\sqrt{n/(n-f(x))}$, which is also illustrated subsequently in~\Cref{fig:policies}. All the best parameters that are greater than one are half-integer in this case (the ones that round up). The other region that is far from the optimum does not contribute much to the expected running time. It is clearly the result of accepting the best mutation offspring if it is better than the parent. In this region, half-integer values (here, the ones that round down) are also typical, but there are also many occurrences of less trivial values, which signifies the necessity of performing proper optimization within the intervals.

It may be tempting to use dynamic programming in a similar way to compute also the best binned policies. However, this time it is not so straightforward. Indeed, if we consider the $i$-th bin and assume that bins $i+1$ up are solved, we can fix some $\lambda$ and compute the expected runtime starting from each fitness value belonging to the $i$-th bin. However, the choice of parameters for the \emph{previous} bins may affect the probability that the algorithm first hits the $i$-th bin at certain fitness value $f$, which would influence the most efficient choice of $\lambda$. As a result, we need to optimize all the parameters \emph{simultaneously}.

For that reason, we use the ideas from~\cite{BuzdalovD21} and employ a numeric optimizer, namely separable CMA-ES~\cite{sep-cma-es}, to find the suitable parameter values for the bins. Based on the insights from~\Cref{sec:exact:landscape}, we represent each parameter $\lambda$ as two variables, $\lambda_i$ and $\lambda_f$, such that $\lambda = \max\{1, \min\{n, \round{\lambda_i \cdot n} + \lambda_f - 0.5\}\}$. These new values are box-constrained, $0 \le \lambda_i, \lambda_f < 1$. This approach makes the optimization landscape smoother, resulting in more stable convergence.

The best results of our optimization attempts for both kinds of policies are presented in~\Cref{table:baseline} together with the obtained parameter values for the bins. Though, strictly speaking, we have no optimality guarantees, we are still pretty confident that the presented values are not just the upper bounds, but rather represent the truly optimal policies up to available precision.

\section{Final Comparison}

\Cref{table:final_comparison} shows the exact expected running times of the best known policies obtained from~\Cref{sec:exact:policies} for both unrestricted (best) and binned versions (best binned), the final configuration found by the dynamic tuning with binning and cascading with the largest number of bins (tuned), and \theorydyn (default). The tunings outperform the default policy for $n \geq 1000$, this is consistent with our empirical evaluations in~\Cref{sec:irace}. However, there is still a rather large gap to the best policies found by our problem-specific customized search methods, which indicates room for further improvement in our tuning approach.

\begin{table}[!t]
\caption{Expected running time of the numerically approximated baseline policies, rounded to two digits after the decimal comma}\label{table:baseline}
\begin{tabular}{r|rr|p{12em}}
$n$ & best & best binned & bin values \\\hline
500 & 2916.94 & 2925.52 & 1, 1, 1, 1, 6.5, 8.5, 11.5, 16.5, 24.5 \\
1000 & 5975.81 & 5994.89 & 1, 1, 1, 1, 6.5, 8.5, 11.5, 16.5, 23.5, 35.5 \\
2000 & 12157.62 & 12197.66 & 1, 1, 1, 1, 6.5, 8.5, 11.5, 16.5, 22.5, 32.5, 49.5
\end{tabular}
\end{table}

\begin{table}[!t]
\caption{Expected runtime of the best approximated policies, the default one and the best dynamic-tuning one}
\label{table:final_comparison}
\begin{tabular}{r|rrrr}
$n$ & best & best binned & default & tuned \\\hline
500 & 2916.94 & 2925.52	& 3224.89 &	3249.09 \\
1000 &	5975.81 &	5994.89 &	6586.67 &	6512.32 \\
2000 &	12157.62 &	12197.66 &	13386.44 &	12703.88 \\
3000 &  18375.48 & 18435.71 & 20128.97	& 19411.0
\end{tabular}
\end{table}

\begin{figure*}[!t]
  \centering
  \includegraphics[width=0.7\textwidth]{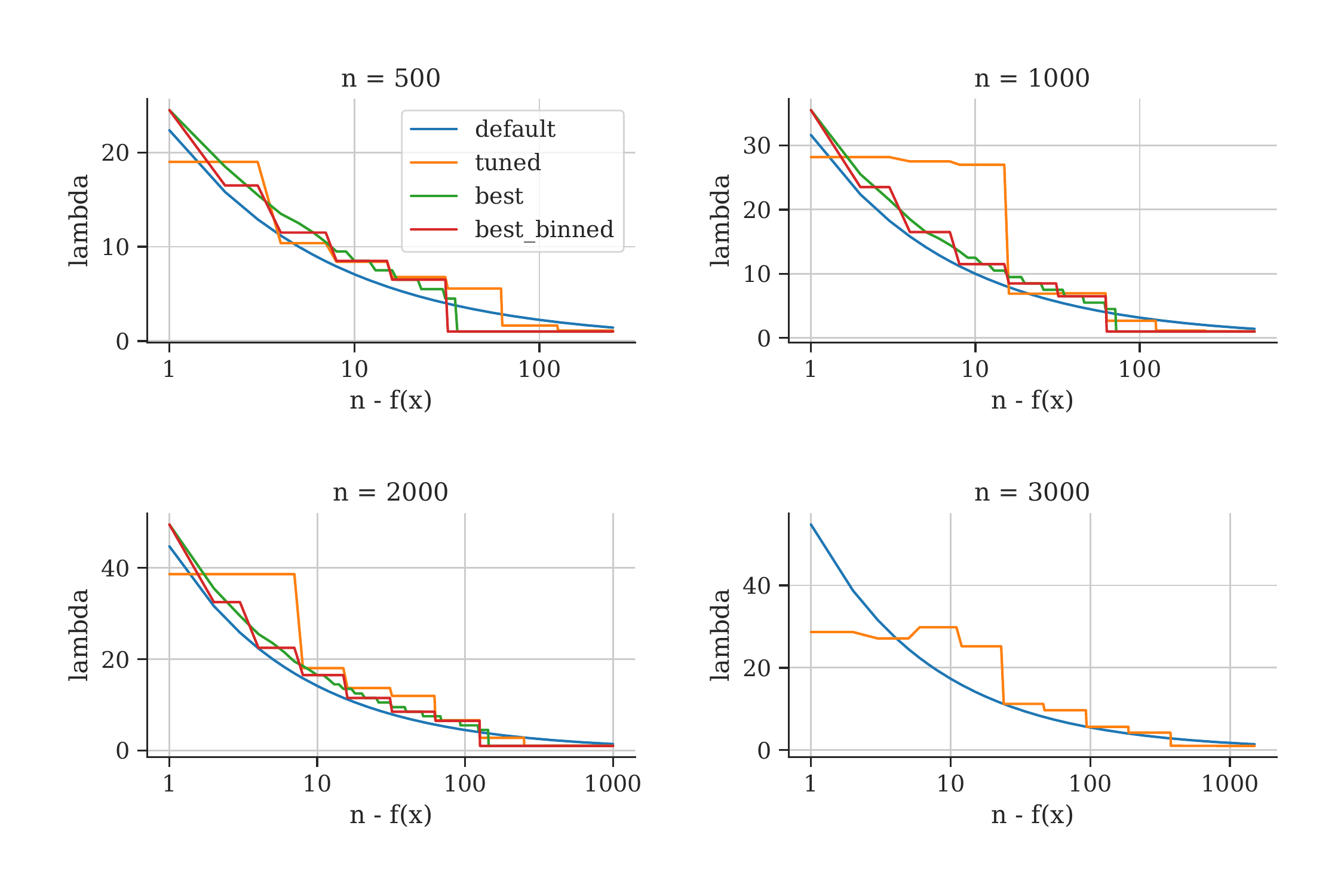}
  \caption{The best approximated policies (\texttt{best} and \texttt{best\_binned}), the default one and the best dynamic one found by the tuning.}
  \Description{The best approximated policies (\texttt{best} and best_binned), the default one and the best dynamic ones found by the tuning.}
  \label{fig:policies}
\end{figure*}

\Cref{fig:policies} shows the policies of those four settings. We only plot $\lambda$ values for $f(x) \geq n/2$ as this is the most interesting region (a random initial solution often has fitness around $n/2$). The $x$-axis is plotted in logarithmic scale of $n-f(x)$ to allow zoom-in effect on the more difficult regions of the objective function. A noticeable observation is that \theorydyn almost always underestimates the best $\lambda$ values in those more difficult areas of the search, and that may be the reason for its worse performance compared to the others. The dynamic tuning, on the other hand, is able to produce policies that are generally close to the best binned policies up until a point. When the fitness is close to the optimal, the tuned policies start diverging from the best \lbd values, indicating the limitation of \irace in detecting the importance of those last few bins in order to tune them properly. 

\section{Insights into Future Work}

The ruggedness of the parameter search landscape highlighted in \Cref{sec:exact:landscape}, and particularly the algorithmic features that cause it, made us believe that there is a rich field for future work considering the design principles of evolutionary algorithms. Rather than just outlining our thoughts, in this section we consider to perform a small series of illustrative experiments that highlight them better.

\subsection{Tunability}

Similarly to the concept of \textit{testability} in engineering disciplines, we want to bring to the focus the concept of \textit{tunability} of evolutionary algorithms. Many design choices (such as whether or not the best mutation offspring is compared to the parent, or whether or not we can sample an offspring which is equal to the parent) look rather innocent and straightforward, and are often done to improve the performance, but they can promptly turn the parameter landscape from a smooth to a complicated rugged surface. For this reason, we think that new evolutionary algorithms should be designed with care about how friendly they are to tuning procedures, that is, with \textit{tunability} in their design. In the context of the \onell, any deterministic rounding of the parameter $\lambda$ to obtain the integer population size $\Lambda$ is likely to be very disruptive to parameter tuning, and some other procedures may be sought for.

\begin{figure}[!t]
    \centering
    \includegraphics[scale=0.85]{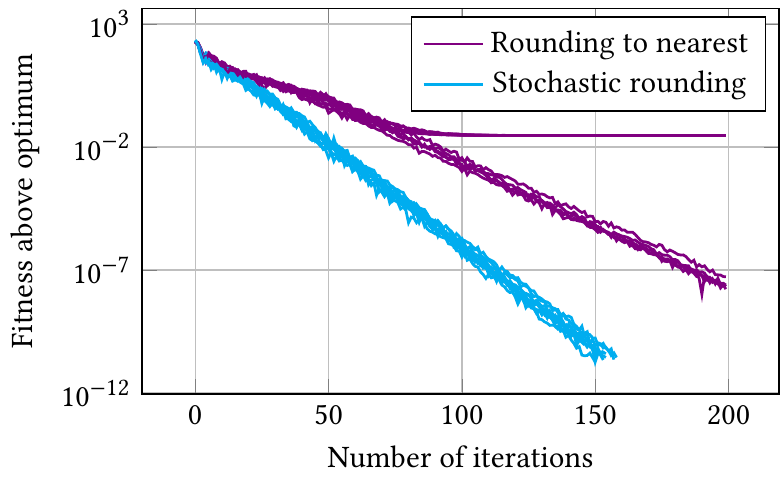}
    \caption{Convergence of parameter optimization for different roundings in \onell in terms of fitness values. 10 runs has been executed for each rounding.}
    \Description{Convergence of parameter optimization for different roundings in terms of fitness values}
    \label{fig:convergence-fitness}
    \vspace{3ex}
    \includegraphics[scale=0.85]{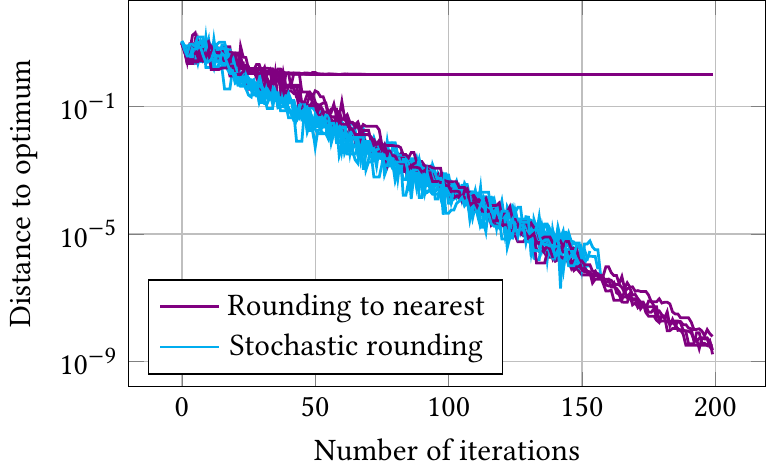}
    \caption{Convergence of parameter optimization for different roundings in \onell in terms of distance to the optimum. 10 runs has been executed for each rounding.}
    \Description{Convergence of parameter optimization for different roundings in terms of distance to the optimum (across multiple runs)}
    \label{fig:convergence-distance}
\end{figure}

\begin{table*}[!t]
    \caption{Optimal policies for $n=100$ and three approaches to coupling parameters in the \onell. All numbers are rounded to four digits after the decimal point}
    \Description{Optimal policies for n=100 and three approaches to coupling parameters}
    \label{tab:separate}
    \centering
    \begin{tabular}{c|c|c|ccccccc}
    \toprule
&&&\multicolumn{7}{c}{Bin}\\
       Coupling             & Runtime  & Param     &  $[0; 50]$ &  $[51; 75]$ &  $[76; 88]$ &  $[89; 94]$ &  $[95; 97]$ &  $[98; 99]$ &  100\\\hline
       Rounding to nearest  & 534.3011 & $\lambda$ & 1.0 & 1.0 & 1.0 & 3.5 & 5.5 & 7.5 & 10.5\\
                            &          & $\Lambda$ & 1   & 1   & 1   & 4   & 6   & 8   & 11 \\\hline
       Stochastic rounding  & 540.7504 & $\lambda$, $\Lambda$ & 1.0 & 1.0 & 1.0 & 4.0 & 6.0 & 8.0688 & 11.2628 \\\hline
       Decoupled parameters & 488.9785 & $\lambda$ & 83.9611 & 1.0 & 1.0 & 1.4670 & 3.0433 & 4.8120 & 6.9308 \\
                            &          & $\Lambda$ & 3 & 1 & 1 & 5 & 8 & 12 & 19\\
    \bottomrule
    \end{tabular}
\end{table*}

To illustrate this point, we compare the version of the \onell studied in this paper with one of the versions from~\cite{bassinB-motor20-oll-choices}, which employs stochastic rounding. 
Here, $\Lambda = \lfloor \lambda \rfloor$ is taken with probability $\lceil \lambda \rceil - \lambda$ and $\Lambda = \lceil \lambda \rceil$ otherwise. We use the problem size $n=100$, the binning policy, and optimize the parameter values for the bins using separable CMA-ES~\cite{sep-cma-es}, but in a straightforward way unlike~\Cref{sec:exact:policies}, that is, each decision variable corresponds to the parameter value. The population size of CMA-ES is 100, and we run optimization for 200 iterations, or until the internal variables of CMA-ES degenerate. \Cref{fig:convergence-fitness} shows the convergence plots with regards to the fitness value, whereas \Cref{fig:convergence-distance} does it for the Euclidean distance to the optimum. Clearly, stochastic rounding shows a much ``cleaner'' convergence without getting stuck in local optima. It also seems somewhat less sensitive with respect to the parameter values around the optimum. We expect that designing algorithms with tunability in mind can bring them similar benefits.

\subsection{Control of multiple parameters}

De-coupling of the parameters of the \oll is interesting from both a theoretical and an empirical point of view. Following the approaches suggested in~\cite{DoerrD18ga,DangD19,bassinB-motor20-oll-choices}, for example, one could explicitly ask to control the population size for the mutation phase, the mutation strength, the population size for the crossover phase, the crossover bias, etc. The method proposed in~\Cref{sec:exact:dp} can be used to model this as well. In a small experiment, we tune the parameter $\lambda$ that controls the mutation strength and the crossover bias, and another parameter~$\Lambda$ that defines the population size for both the mutation and the crossover phases. Just as above, we use $n=100$, the binning policy, and optimize the parameters using separable CMA-ES. 

\Cref{tab:separate} shows the results and compares them with the previously discussed configuration. While rounding to the nearest integer performs slightly better than stochastic rounding in terms of the runtime of the \onell, it appears to be that it happens because of $\lambda$ and $\Lambda$ being coupled, and the decoupled version wins by a large margin. Note that the decoupled version favors much larger population sizes $\Lambda$ and much smaller variation strengths $\lambda$ towards the optimum. However, it benefits from small population sizes and large mutation strength of almost $0.84$ in the first bin, which can essentially flip all the bits at once. Note that it is impossible to benefit from this behavior with tightly coupled parameters.  

\balance
\section{Conclusion and Outlook}
\label{sec:conclusion}

In this work, we have evaluated some basic properties of population size control of the \oll solving \onemax as a DAC benchmark. We empirically evaluated the effectiveness of an automated algorithm configuration approach using \irace as a DAC solving method. Our results suggest that such approach has the potential of learning well-performing parameter control policies, but it has to be implemented carefully. More concretely, a naive implementation would completely fail to get anywhere close to a good policy on our benchmark, but an improved version with binning and cascading can help to boost the performance considerably. 

Our work introduces an alternative way to train control policies in DAC settings, a problem that was previously studied predominantly from the viewpoint of reinforcement learning (RL)~\cite{ManuelGECCO2019DDQN,Biedenkapp22thesis,BiedenkappDKHD22GECCO} or (in numerical optimization) using exploratory landscape analysis~\cite{AnjaPPSN2022}. We believe that each one of them has different strengths and weaknesses. Having a rich collection of DAC benchmarks with well-understood ground-truth information allows proper investigation of the potential and the limitations of each family of DAC approaches. Our work contributes to such benchmarks. 

We did not compare our results with the RL-based approaches in~\cite{ManuelGECCO2019DDQN,BiedenkappDKHD22GECCO} as the RL algorithms used in those work (DDQN~\cite{van2016deep}) assume a discrete action space where every action value is treated as categorical, while in our setting the parameter being controlled (\lbd) is numerical. We did, however, conduct some preliminary experiments with PPO~\cite{SchulmanWDRK17PPO}, a well-known RL approach that works with both discrete and continuous action spaces, but the results were rather disappointing. This may not be too surprising, as deep-RL algorithms are often known to be non-trivial to use and various design choices may have strong impact on their performance~\cite{henderson2018deep,andrychowicz2021matters}. A thorough study on deep-RL for the benchmark is beyond the scope of this present work. We leave it as a future research question.

From the perspective of analysis and design of evolutionary computation methods, our numerical approximations of the optimal control policies raise a number of interesting questions. It appears that the design choice in the considered version of the \oll, namely deterministic rounding of a real-valued parameter to obtain population size, led to a rugged landscape of the parameter search space, which not only complicated the numerical minimization, but probably had an effect on the application of \irace. A rather simple change, switching to stochastic rounding, made the parameter tuning problem much simpler, but so did optimizing two parameters (the original $\lambda$ and the population size $\Lambda$) separately. As a result of these exercises, we propose considering tunability of evolutionary algorithms when designing them, and note that sometimes tuning more parameters is easier. With this in hand, we can also construct new DAC benchmarks that require to control not only one, but several parameters at the same time~\cite{xue2022multi}~-- a largely under-explored challenge~\cite{KarafotiasHE15,Adriansen2022DACjournal}.

\begin{acks}
Nguyen Dang is a Leverhulme Early Career Fellow. 
We used the Cirrus UK National Tier-2 HPC Service at EPCC (\url{http://www.cirrus.ac.uk}) funded by the University of Edinburgh and EPSRC (EP/P020267/1). Deyao Chen is supported by the St Andrews Research Internship Scheme (StARIS). 
We furthermore acknowledge financial support by ANR-22-ERCS-0003-01 project VARIATION. 
\end{acks}

\bibliographystyle{ACM-Reference-Format}
\bibliography{references}


\begin{thebibliography}{54}


\ifx \showCODEN    \undefined \def \showCODEN     #1{\unskip}     \fi
\ifx \showDOI      \undefined \def \showDOI       #1{#1}\fi
\ifx \showISBNx    \undefined \def \showISBNx     #1{\unskip}     \fi
\ifx \showISBNxiii \undefined \def \showISBNxiii  #1{\unskip}     \fi
\ifx \showISSN     \undefined \def \showISSN      #1{\unskip}     \fi
\ifx \showLCCN     \undefined \def \showLCCN      #1{\unskip}     \fi
\ifx \shownote     \undefined \def \shownote      #1{#1}          \fi
\ifx \showarticletitle \undefined \def \showarticletitle #1{#1}   \fi
\ifx \showURL      \undefined \def \showURL       {\relax}        \fi
\providecommand\bibfield[2]{#2}
\providecommand\bibinfo[2]{#2}
\providecommand\natexlab[1]{#1}
\providecommand\showeprint[2][]{arXiv:#2}

\bibitem[Adriaensen et~al\mbox{.}(2022)]%
        {Adriansen2022DACjournal}
\bibfield{author}{\bibinfo{person}{Steven Adriaensen},
  \bibinfo{person}{Andr{\'{e}} Biedenkapp}, \bibinfo{person}{Gresa Shala},
  \bibinfo{person}{Noor Awad}, \bibinfo{person}{Theresa Eimer},
  \bibinfo{person}{Marius Lindauer}, {and} \bibinfo{person}{Frank Hutter}.}
  \bibinfo{year}{2022}\natexlab{}.
\newblock \showarticletitle{Automated Dynamic Algorithm Configuration}.
\newblock \bibinfo{journal}{\emph{Journal of Artificial Intelligence Research}}
   \bibinfo{volume}{75} (\bibinfo{year}{2022}), \bibinfo{pages}{1633--1699}.
\newblock
\urldef\tempurl%
\url{https://doi.org/10.1613/jair.1.13922}
\showDOI{\tempurl}


\bibitem[Andrychowicz et~al\mbox{.}(2021)]%
        {andrychowicz2021matters}
\bibfield{author}{\bibinfo{person}{Marcin Andrychowicz}, \bibinfo{person}{Anton
  Raichuk}, \bibinfo{person}{Piotr Sta{\'n}czyk}, \bibinfo{person}{Manu
  Orsini}, \bibinfo{person}{Sertan Girgin}, \bibinfo{person}{Rapha{\"e}l
  Marinier}, \bibinfo{person}{Leonard Hussenot}, \bibinfo{person}{Matthieu
  Geist}, \bibinfo{person}{Olivier Pietquin}, \bibinfo{person}{Marcin
  Michalski}, {et~al\mbox{.}}} \bibinfo{year}{2021}\natexlab{}.
\newblock \showarticletitle{What matters for on-policy deep actor-critic
  methods? a large-scale study}. In \bibinfo{booktitle}{\emph{International
  conference on learning representations}}.
\newblock


\bibitem[Antipov et~al\mbox{.}(2022a)]%
        {AntipovBD22}
\bibfield{author}{\bibinfo{person}{Denis Antipov}, \bibinfo{person}{Maxim
  Buzdalov}, {and} \bibinfo{person}{Benjamin Doerr}.}
  \bibinfo{year}{2022}\natexlab{a}.
\newblock \showarticletitle{Fast Mutation in Crossover-Based Algorithms}.
\newblock \bibinfo{journal}{\emph{Algorithmica}} \bibinfo{volume}{84},
  \bibinfo{number}{6} (\bibinfo{year}{2022}), \bibinfo{pages}{1724--1761}.
\newblock
\urldef\tempurl%
\url{https://doi.org/10.1007/s00453-022-00957-5}
\showDOI{\tempurl}


\bibitem[Antipov et~al\mbox{.}(2019)]%
        {AntipovDK19LO}
\bibfield{author}{\bibinfo{person}{Denis Antipov}, \bibinfo{person}{Benjamin
  Doerr}, {and} \bibinfo{person}{Vitalii Karavaev}.}
  \bibinfo{year}{2019}\natexlab{}.
\newblock \showarticletitle{A tight runtime analysis for the
  $(1+(\lambda,\lambda))$~{GA} on {L}eading{O}nes}. In
  \bibinfo{booktitle}{\emph{Proc. of Foundations of Genetic Algorithms
  (FOGA'19)}}. \bibinfo{publisher}{ACM}, \bibinfo{pages}{169--182}.
\newblock
\urldef\tempurl%
\url{https://doi.org/10.1145/3299904.3340317}
\showDOI{\tempurl}


\bibitem[Antipov et~al\mbox{.}(2022b)]%
        {AntipovDK22}
\bibfield{author}{\bibinfo{person}{Denis Antipov}, \bibinfo{person}{Benjamin
  Doerr}, {and} \bibinfo{person}{Vitalii Karavaev}.}
  \bibinfo{year}{2022}\natexlab{b}.
\newblock \showarticletitle{A Rigorous Runtime Analysis of the {(1} +
  ({\(\lambda\)} , {\(\lambda\)} {))} {GA} on Jump Functions}.
\newblock \bibinfo{journal}{\emph{Algorithmica}} \bibinfo{volume}{84},
  \bibinfo{number}{6} (\bibinfo{year}{2022}), \bibinfo{pages}{1573--1602}.
\newblock
\urldef\tempurl%
\url{https://doi.org/10.1007/s00453-021-00907-7}
\showDOI{\tempurl}


\bibitem[Antonov et~al\mbox{.}(2021)]%
        {AntonovBBD21}
\bibfield{author}{\bibinfo{person}{Kirill Antonov}, \bibinfo{person}{Maxim
  Buzdalov}, \bibinfo{person}{Arina Buzdalova}, {and} \bibinfo{person}{Carola
  Doerr}.} \bibinfo{year}{2021}\natexlab{}.
\newblock \showarticletitle{Blending Dynamic Programming with Monte Carlo
  Simulation for Bounding the Running Time of Evolutionary Algorithms}. In
  \bibinfo{booktitle}{\emph{Proc. of {IEEE} Congress on Evolutionary
  Computation (CEC'21)}}. \bibinfo{publisher}{IEEE}, \bibinfo{pages}{878--885}.
\newblock
\urldef\tempurl%
\url{https://doi.org/10.1109/CEC45853.2021.9504775}
\showDOI{\tempurl}
\newblock
\shownote{Free version available at \url{https://arxiv.org/abs/2102.11461}}.


\bibitem[Bartz{-}Beielstein et~al\mbox{.}(2020)]%
        {TBB20benchmarking}
\bibfield{author}{\bibinfo{person}{Thomas Bartz{-}Beielstein},
  \bibinfo{person}{Carola Doerr}, \bibinfo{person}{Jakob Bossek},
  \bibinfo{person}{Sowmya Chandrasekaran}, \bibinfo{person}{Tome Eftimov},
  \bibinfo{person}{Andreas Fischbach}, \bibinfo{person}{Pascal Kerschke},
  \bibinfo{person}{Manuel L{\'{o}}pez{-}Ib{\'{a}}{\~{n}}ez},
  \bibinfo{person}{Katherine~M. Malan}, \bibinfo{person}{Jason~H. Moore},
  \bibinfo{person}{Boris Naujoks}, \bibinfo{person}{Patryk Orzechowski},
  \bibinfo{person}{Vanessa Volz}, \bibinfo{person}{Markus Wagner}, {and}
  \bibinfo{person}{Thomas Weise}.} \bibinfo{year}{2020}\natexlab{}.
\newblock \showarticletitle{Benchmarking in Optimization: Best Practice and
  Open Issues}.
\newblock \bibinfo{journal}{\emph{CoRR}}  \bibinfo{volume}{abs/2007.03488}
  (\bibinfo{year}{2020}).
\newblock
\showeprint[arxiv]{2007.03488}
\urldef\tempurl%
\url{https://arxiv.org/abs/2007.03488}
\showURL{%
\tempurl}


\bibitem[Bassin and Buzdalov(2020)]%
        {bassinB-motor20-oll-choices}
\bibfield{author}{\bibinfo{person}{Anton Bassin} {and} \bibinfo{person}{Maxim
  Buzdalov}.} \bibinfo{year}{2020}\natexlab{}.
\newblock \showarticletitle{An Experimental Study of Operator Choices in the
  {$(1+(\lambda,\lambda))$}~Genetic Algorithm}.
\newblock In \bibinfo{booktitle}{\emph{Proceedings of the International
  Conference on Mathematical Optimization Theory and Operations Research}}.
  Number 1275 in \bibinfo{series}{Communications in Computer and Information
  Science}. \bibinfo{pages}{320--335}.
\newblock
\urldef\tempurl%
\url{https://doi.org/10.1007/978-3-030-58657-7_26}
\showDOI{\tempurl}


\bibitem[Biedenkapp(2022)]%
        {Biedenkapp22thesis}
\bibfield{author}{\bibinfo{person}{Andr{\'{e}} Biedenkapp}.}
  \bibinfo{year}{2022}\natexlab{}.
\newblock \emph{\bibinfo{title}{Dynamic algorithm configuration by
  reinforcement learning}}.
\newblock \bibinfo{thesistype}{Ph.\,D. Dissertation}.
  \bibinfo{school}{University of Freiburg, Germany}.
\newblock
\urldef\tempurl%
\url{https://freidok.uni-freiburg.de/data/230869}
\showURL{%
\tempurl}


\bibitem[Biedenkapp et~al\mbox{.}(2020)]%
        {BiedenkappBEHL20DACECAI}
\bibfield{author}{\bibinfo{person}{Andr{\'{e}} Biedenkapp},
  \bibinfo{person}{H.~Furkan Bozkurt}, \bibinfo{person}{Theresa Eimer},
  \bibinfo{person}{Frank Hutter}, {and} \bibinfo{person}{Marius Lindauer}.}
  \bibinfo{year}{2020}\natexlab{}.
\newblock \showarticletitle{Dynamic Algorithm Configuration: Foundation of a
  New Meta-Algorithmic Framework}. In \bibinfo{booktitle}{\emph{Proc. of
  European Conference on Artificial Intelligence (ECAI'20)}}
  \emph{(\bibinfo{series}{Frontiers in Artificial Intelligence and
  Applications}, Vol.~\bibinfo{volume}{325})}. \bibinfo{publisher}{{IOS}
  Press}, \bibinfo{pages}{427--434}.
\newblock
\urldef\tempurl%
\url{https://doi.org/10.3233/FAIA200122}
\showDOI{\tempurl}


\bibitem[Biedenkapp et~al\mbox{.}(2022)]%
        {BiedenkappDKHD22GECCO}
\bibfield{author}{\bibinfo{person}{Andr{\'{e}} Biedenkapp},
  \bibinfo{person}{Nguyen Dang}, \bibinfo{person}{Martin~S. Krejca},
  \bibinfo{person}{Frank Hutter}, {and} \bibinfo{person}{Carola Doerr}.}
  \bibinfo{year}{2022}\natexlab{}.
\newblock \showarticletitle{Theory-inspired parameter control benchmarks for
  dynamic algorithm configuration}. In \bibinfo{booktitle}{\emph{Proc. of
  Genetic and Evolutionary Computation Conference (GECCO)}}.
  \bibinfo{publisher}{ACM}, \bibinfo{pages}{766--775}.
\newblock
\urldef\tempurl%
\url{https://doi.org/10.1145/3512290.3528846}
\showDOI{\tempurl}


\bibitem[Burke et~al\mbox{.}(2013)]%
        {BurkeGHKOOQ13}
\bibfield{author}{\bibinfo{person}{Edmund~K. Burke}, \bibinfo{person}{Michel
  Gendreau}, \bibinfo{person}{Matthew~R. Hyde}, \bibinfo{person}{Graham
  Kendall}, \bibinfo{person}{Gabriela Ochoa}, \bibinfo{person}{Ender
  {\"{O}}zcan}, {and} \bibinfo{person}{Rong Qu}.}
  \bibinfo{year}{2013}\natexlab{}.
\newblock \showarticletitle{Hyper-heuristics: a survey of the state of the
  art}.
\newblock \bibinfo{journal}{\emph{J. Oper. Res. Soc.}} \bibinfo{volume}{64},
  \bibinfo{number}{12} (\bibinfo{year}{2013}), \bibinfo{pages}{1695--1724}.
\newblock
\urldef\tempurl%
\url{https://doi.org/10.1057/jors.2013.71}
\showDOI{\tempurl}


\bibitem[Buskulic and Doerr(2021)]%
        {BuskulicD21}
\bibfield{author}{\bibinfo{person}{Nathan Buskulic} {and}
  \bibinfo{person}{Carola Doerr}.} \bibinfo{year}{2021}\natexlab{}.
\newblock \showarticletitle{Maximizing Drift Is Not Optimal for Solving
  OneMax}.
\newblock \bibinfo{journal}{\emph{Evol. Comput.}} \bibinfo{volume}{29},
  \bibinfo{number}{4} (\bibinfo{year}{2021}), \bibinfo{pages}{521--541}.
\newblock
\urldef\tempurl%
\url{https://doi.org/10.1162/evco\_a\_00290}
\showDOI{\tempurl}


\bibitem[Buzdalov and Doerr(2017)]%
        {BuzdalovD17}
\bibfield{author}{\bibinfo{person}{Maxim Buzdalov} {and}
  \bibinfo{person}{Benjamin Doerr}.} \bibinfo{year}{2017}\natexlab{}.
\newblock \showarticletitle{Runtime Analysis of the $(1+(\lambda,\lambda))$
  {Genetic Algorithm} on Random Satisfiable {3-CNF} Formulas}. In
  \bibinfo{booktitle}{\emph{Proc. of Genetic and Evolutionary Computation
  Conference (GECCO'17)}}. \bibinfo{publisher}{ACM},
  \bibinfo{pages}{1343--1350}.
\newblock


\bibitem[Buzdalov and Doerr(2020)]%
        {BuzdalovD20}
\bibfield{author}{\bibinfo{person}{Maxim Buzdalov} {and}
  \bibinfo{person}{Carola Doerr}.} \bibinfo{year}{2020}\natexlab{}.
\newblock \showarticletitle{Optimal Mutation Rates for the $(1+\lambda)$ {EA}
  on {O}ne{M}ax}. In \bibinfo{booktitle}{\emph{Proc. of Parallel Problem
  Solving from Nature (PPSN'20)}} \emph{(\bibinfo{series}{LNCS},
  Vol.~\bibinfo{volume}{12270})}. \bibinfo{publisher}{Springer},
  \bibinfo{pages}{574--587}.
\newblock
\urldef\tempurl%
\url{https://doi.org/10.1007/978-3-030-58115-2\_40}
\showDOI{\tempurl}


\bibitem[Buzdalov and Doerr(2021)]%
        {BuzdalovD21}
\bibfield{author}{\bibinfo{person}{Maxim Buzdalov} {and}
  \bibinfo{person}{Carola Doerr}.} \bibinfo{year}{2021}\natexlab{}.
\newblock \showarticletitle{Optimal static mutation strength distributions for
  the $(1+\lambda)$ evolutionary algorithm on OneMax}. In
  \bibinfo{booktitle}{\emph{Proc. of Genetic and Evolutionary Computation
  Conference (GECCO'21)}}. \bibinfo{publisher}{ACM}, \bibinfo{pages}{660--668}.
\newblock
\urldef\tempurl%
\url{https://doi.org/10.1145/3449639.3459389}
\showDOI{\tempurl}


\bibitem[C{\'a}ceres et~al\mbox{.}(2017)]%
        {caceres2017experimental}
\bibfield{author}{\bibinfo{person}{Leslie~P{\'e}rez C{\'a}ceres},
  \bibinfo{person}{Manuel L{\'o}pez-Ib{\'a}{\~n}ez}, \bibinfo{person}{Holger
  Hoos}, {and} \bibinfo{person}{Thomas St{\"u}tzle}.}
  \bibinfo{year}{2017}\natexlab{}.
\newblock \showarticletitle{An experimental study of adaptive capping in
  irace}. In \bibinfo{booktitle}{\emph{Learning and Intelligent Optimization:
  11th International Conference, LION 11, Nizhny Novgorod, Russia, June 19-21,
  2017, Revised Selected Papers}}. Springer, \bibinfo{pages}{235--250}.
\newblock


\bibitem[Chen et~al\mbox{.}(2023)]%
        {data}
\bibfield{author}{\bibinfo{person}{Deyao Chen}, \bibinfo{person}{Maxim
  Buzdalov}, \bibinfo{person}{Carola Doerr}, {and} \bibinfo{person}{Nguyen
  Dang}.} \bibinfo{year}{2023}\natexlab{}.
\newblock \bibinfo{title}{Code and data repository of this paper}.
\newblock \bibinfo{howpublished}{\url{https://github.com/de0ch/OLL}}.
\newblock


\bibitem[Dang and Doerr(2019)]%
        {DangD19}
\bibfield{author}{\bibinfo{person}{Nguyen Dang} {and} \bibinfo{person}{Carola
  Doerr}.} \bibinfo{year}{2019}\natexlab{}.
\newblock \showarticletitle{Hyper-parameter tuning for the
  $(1+(\lambda,\lambda))$~{GA}}. In \bibinfo{booktitle}{\emph{Proc. of Genetic
  and Evolutionary Computation Conference (GECO'19)}}.
  \bibinfo{publisher}{ACM}, \bibinfo{pages}{889--897}.
\newblock
\urldef\tempurl%
\url{https://doi.org/10.1145/3321707.3321725}
\showDOI{\tempurl}


\bibitem[de~Souza et~al\mbox{.}(2022)]%
        {SouzaRL22iraceCapping}
\bibfield{author}{\bibinfo{person}{Marcelo de Souza}, \bibinfo{person}{Marcus
  Ritt}, {and} \bibinfo{person}{Manuel L{\'{o}}pez{-}Ib{\'{a}}{\~{n}}ez}.}
  \bibinfo{year}{2022}\natexlab{}.
\newblock \showarticletitle{Capping methods for the automatic configuration of
  optimization algorithms}.
\newblock \bibinfo{journal}{\emph{Comput. Oper. Res.}}  \bibinfo{volume}{139}
  (\bibinfo{year}{2022}), \bibinfo{pages}{105615}.
\newblock
\urldef\tempurl%
\url{https://doi.org/10.1016/j.cor.2021.105615}
\showDOI{\tempurl}


\bibitem[Doerr(2019)]%
        {Doerr19domi}
\bibfield{author}{\bibinfo{person}{Benjamin Doerr}.}
  \bibinfo{year}{2019}\natexlab{}.
\newblock \showarticletitle{Analyzing randomized search heuristics via
  stochastic domination}.
\newblock \bibinfo{journal}{\emph{Theoretical Computer Science}}
  \bibinfo{volume}{773} (\bibinfo{year}{2019}), \bibinfo{pages}{115--137}.
\newblock
\urldef\tempurl%
\url{https://doi.org/10.1016/j.tcs.2018.09.024}
\showDOI{\tempurl}


\bibitem[Doerr and Doerr(2018)]%
        {DoerrD18ga}
\bibfield{author}{\bibinfo{person}{Benjamin Doerr} {and}
  \bibinfo{person}{Carola Doerr}.} \bibinfo{year}{2018}\natexlab{}.
\newblock \showarticletitle{Optimal Static and Self-Adjusting Parameter Choices
  for the (1+($\lambda$,$\lambda$)) Genetic Algorithm}.
\newblock \bibinfo{journal}{\emph{Algorithmica}}  \bibinfo{volume}{80}
  (\bibinfo{year}{2018}), \bibinfo{pages}{1658--1709}.
\newblock
\urldef\tempurl%
\url{https://doi.org/10.1007/s00453-017-0354-9}
\showDOI{\tempurl}


\bibitem[Doerr and Doerr(2020)]%
        {DoerrD18chapter}
\bibfield{author}{\bibinfo{person}{Benjamin Doerr} {and}
  \bibinfo{person}{Carola Doerr}.} \bibinfo{year}{2020}\natexlab{}.
\newblock \showarticletitle{Theory of Parameter Control Mechanisms for Discrete
  Black-Box Optimization: Provable Performance Gains Through Dynamic Parameter
  Choices}.
\newblock In \bibinfo{booktitle}{\emph{Theory of Evolutionary Computation:
  Recent Developments in Discrete Optimization}}.
  \bibinfo{publisher}{Springer}, \bibinfo{pages}{271--321}.
\newblock


\bibitem[Doerr et~al\mbox{.}(2015)]%
        {DoerrDE15}
\bibfield{author}{\bibinfo{person}{Benjamin Doerr}, \bibinfo{person}{Carola
  Doerr}, {and} \bibinfo{person}{Franziska Ebel}.}
  \bibinfo{year}{2015}\natexlab{}.
\newblock \showarticletitle{From black-box complexity to designing new genetic
  algorithms}.
\newblock \bibinfo{journal}{\emph{Theoretical Computer Science}}
  \bibinfo{volume}{567} (\bibinfo{year}{2015}), \bibinfo{pages}{87 -- 104}.
\newblock


\bibitem[Doerr et~al\mbox{.}(2017)]%
        {fastGA}
\bibfield{author}{\bibinfo{person}{Benjamin Doerr}, \bibinfo{person}{Huu~Phuoc
  Le}, \bibinfo{person}{R{\'{e}}gis Makhmara}, {and} \bibinfo{person}{Ta~Duy
  Nguyen}.} \bibinfo{year}{2017}\natexlab{}.
\newblock \showarticletitle{Fast genetic algorithms}. In
  \bibinfo{booktitle}{\emph{Proc. of Genetic and Evolutionary Computation
  Conference (GECCO'17)}}. \bibinfo{publisher}{ACM}, \bibinfo{pages}{777--784}.
\newblock
\urldef\tempurl%
\url{https://doi.org/10.1145/3071178.3071301}
\showDOI{\tempurl}


\bibitem[Doerr and Winzen(2014)]%
        {DoerrW14memory}
\bibfield{author}{\bibinfo{person}{Benjamin Doerr} {and}
  \bibinfo{person}{Carola Winzen}.} \bibinfo{year}{2014}\natexlab{}.
\newblock \showarticletitle{Playing {M}astermind with Constant-Size Memory}.
\newblock \bibinfo{journal}{\emph{Theory of Computing Systems}}
  \bibinfo{volume}{55} (\bibinfo{year}{2014}), \bibinfo{pages}{658--684}.
\newblock


\bibitem[Doerr(2020)]%
        {Doerr20chapter}
\bibfield{author}{\bibinfo{person}{Carola Doerr}.}
  \bibinfo{year}{2020}\natexlab{}.
\newblock \showarticletitle{Complexity Theory for Black-Box Optimization
  Heuristics}.
\newblock In \bibinfo{booktitle}{\emph{Theory of Evolutionary Computation:
  Recent Developments in Discrete Optimization}}.
  \bibinfo{publisher}{Springer}, \bibinfo{pages}{133--212}.
\newblock


\bibitem[Doerr and Wagner(2018)]%
        {DoerrW18}
\bibfield{author}{\bibinfo{person}{Carola Doerr} {and} \bibinfo{person}{Markus
  Wagner}.} \bibinfo{year}{2018}\natexlab{}.
\newblock \showarticletitle{Simple on-the-fly parameter selection mechanisms
  for two classical discrete black-box optimization benchmark problems}. In
  \bibinfo{booktitle}{\emph{Proc. of Genetic and Evolutionary Computation
  Conference (GECCO'18)}}. \bibinfo{publisher}{ACM}, \bibinfo{pages}{943--950}.
\newblock
\urldef\tempurl%
\url{https://doi.org/10.1145/3205455.3205560}
\showDOI{\tempurl}


\bibitem[Eiben et~al\mbox{.}(1999)]%
        {EibenHM99}
\bibfield{author}{\bibinfo{person}{Agoston~Endre Eiben},
  \bibinfo{person}{Robert Hinterding}, {and} \bibinfo{person}{Zbigniew
  Michalewicz}.} \bibinfo{year}{1999}\natexlab{}.
\newblock \showarticletitle{Parameter control in evolutionary algorithms}.
\newblock \bibinfo{journal}{\emph{IEEE Transactions on Evolutionary
  Computation}}  \bibinfo{volume}{3} (\bibinfo{year}{1999}),
  \bibinfo{pages}{124--141}.
\newblock


\bibitem[Eimer et~al\mbox{.}(2021)]%
        {DACBench}
\bibfield{author}{\bibinfo{person}{Theresa Eimer}, \bibinfo{person}{Andr{\'{e}}
  Biedenkapp}, \bibinfo{person}{Maximilian Reimer}, \bibinfo{person}{Steven
  Adriaensen}, \bibinfo{person}{Frank Hutter}, {and} \bibinfo{person}{Marius
  Lindauer}.} \bibinfo{year}{2021}\natexlab{}.
\newblock \showarticletitle{DACBench: {A} Benchmark Library for Dynamic
  Algorithm Configuration}. In \bibinfo{booktitle}{\emph{Proc. of International
  Joint Conference on Artificial Intelligence (IJCAI'21)}}.
  \bibinfo{publisher}{ijcai.org}, \bibinfo{pages}{1668--1674}.
\newblock
\urldef\tempurl%
\url{https://doi.org/10.24963/ijcai.2021/230}
\showDOI{\tempurl}


\bibitem[Fialho et~al\mbox{.}(2010)]%
        {FialhoCSS10journal}
\bibfield{author}{\bibinfo{person}{{\'{A}}lvaro Fialho},
  \bibinfo{person}{Lu{\'{\i}}s~Da Costa}, \bibinfo{person}{Marc Schoenauer},
  {and} \bibinfo{person}{Mich{\`{e}}le Sebag}.}
  \bibinfo{year}{2010}\natexlab{}.
\newblock \showarticletitle{Analyzing bandit-based adaptive operator selection
  mechanisms}.
\newblock \bibinfo{journal}{\emph{Annals of Mathematics and Artificial
  Intelligence}}  \bibinfo{volume}{60} (\bibinfo{year}{2010}),
  \bibinfo{pages}{25--64}.
\newblock
\urldef\tempurl%
\url{https://doi.org/10.1007/s10472-010-9213-y}
\showURL{%
\tempurl}


\bibitem[Goldman and Punch(2015)]%
        {GoldmanP15}
\bibfield{author}{\bibinfo{person}{Brian~W. Goldman} {and}
  \bibinfo{person}{William~F. Punch}.} \bibinfo{year}{2015}\natexlab{}.
\newblock \showarticletitle{Fast and Efficient Black Box Optimization Using the
  Parameter-less Population Pyramid}.
\newblock \bibinfo{journal}{\emph{Evolutionary Computation}}
  \bibinfo{volume}{23} (\bibinfo{year}{2015}), \bibinfo{pages}{451--479}.
\newblock


\bibitem[Hansen et~al\mbox{.}(2020)]%
        {hansen2020cocoJournal}
\bibfield{author}{\bibinfo{person}{Nikolaus Hansen}, \bibinfo{person}{Anne
  Auger}, \bibinfo{person}{Raymond Ros}, \bibinfo{person}{Olaf Mersmann},
  \bibinfo{person}{Tea Tu{\v s}ar}, {and} \bibinfo{person}{Dimo Brockhoff}.}
  \bibinfo{year}{2020}\natexlab{}.
\newblock \showarticletitle{{COCO:} a platform for comparing continuous
  optimizers in a black-box setting}.
\newblock \bibinfo{journal}{\emph{Optimization Methods and Software}}
  (\bibinfo{year}{2020}), \bibinfo{pages}{1--31}.
\newblock
\urldef\tempurl%
\url{https://doi.org/10.1080/10556788.2020.1808977}
\showDOI{\tempurl}
\showeprint{https://doi.org/10.1080/10556788.2020.1808977}


\bibitem[Henderson et~al\mbox{.}(2018)]%
        {henderson2018deep}
\bibfield{author}{\bibinfo{person}{Peter Henderson}, \bibinfo{person}{Riashat
  Islam}, \bibinfo{person}{Philip Bachman}, \bibinfo{person}{Joelle Pineau},
  \bibinfo{person}{Doina Precup}, {and} \bibinfo{person}{David Meger}.}
  \bibinfo{year}{2018}\natexlab{}.
\newblock \showarticletitle{Deep reinforcement learning that matters}. In
  \bibinfo{booktitle}{\emph{Proceedings of the AAAI conference on artificial
  intelligence}}, Vol.~\bibinfo{volume}{32}.
\newblock


\bibitem[Hutter et~al\mbox{.}(2014)]%
        {hutter2014aclib}
\bibfield{author}{\bibinfo{person}{Frank Hutter}, \bibinfo{person}{Manuel
  L{\'o}pez-Ib{\'a}nez}, \bibinfo{person}{Chris Fawcett},
  \bibinfo{person}{Marius Lindauer}, \bibinfo{person}{Holger~H Hoos},
  \bibinfo{person}{Kevin Leyton-Brown}, {and} \bibinfo{person}{Thomas
  St{\"u}tzle}.} \bibinfo{year}{2014}\natexlab{}.
\newblock \showarticletitle{AClib: A benchmark library for algorithm
  configuration}. In \bibinfo{booktitle}{\emph{Learning and Intelligent
  Optimization: 8th International Conference, Lion 8, Gainesville, FL, USA,
  February 16-21, 2014. Revised Selected Papers 8}}. Springer,
  \bibinfo{pages}{36--40}.
\newblock


\bibitem[Karafotias et~al\mbox{.}(2015)]%
        {KarafotiasHE15}
\bibfield{author}{\bibinfo{person}{Giorgos Karafotias}, \bibinfo{person}{Mark
  Hoogendoorn}, {and} \bibinfo{person}{A.E. Eiben}.}
  \bibinfo{year}{2015}\natexlab{}.
\newblock \showarticletitle{Parameter Control in Evolutionary Algorithms:
  Trends and Challenges}.
\newblock \bibinfo{journal}{\emph{IEEE Transactions on Evolutionary
  Computation}}  \bibinfo{volume}{19} (\bibinfo{year}{2015}),
  \bibinfo{pages}{167--187}.
\newblock


\bibitem[Kern et~al\mbox{.}(2004)]%
        {KernMHBOK04}
\bibfield{author}{\bibinfo{person}{Stefan Kern}, \bibinfo{person}{Sibylle~D.
  M{\"{u}}ller}, \bibinfo{person}{Nikolaus Hansen}, \bibinfo{person}{Dirk
  B{\"{u}}che}, \bibinfo{person}{Jiri Ocenasek}, {and} \bibinfo{person}{Petros
  Koumoutsakos}.} \bibinfo{year}{2004}\natexlab{}.
\newblock \showarticletitle{Learning probability distributions in continuous
  evolutionary algorithms - a comparative review}.
\newblock \bibinfo{journal}{\emph{Natural Computing}}  \bibinfo{volume}{3}
  (\bibinfo{year}{2004}), \bibinfo{pages}{77--112}.
\newblock


\bibitem[Kostovska et~al\mbox{.}(2022)]%
        {AnjaPPSN2022}
\bibfield{author}{\bibinfo{person}{Ana Kostovska}, \bibinfo{person}{Anja
  Jankovic}, \bibinfo{person}{Diederick Vermetten}, \bibinfo{person}{Jacob de
  Nobel}, \bibinfo{person}{Hao Wang}, \bibinfo{person}{Tome Eftimov}, {and}
  \bibinfo{person}{Carola Doerr}.} \bibinfo{year}{2022}\natexlab{}.
\newblock \showarticletitle{Per-run Algorithm Selection with Warm-starting
  using Trajectory-based Features}. In \bibinfo{booktitle}{\emph{Parallel
  Problem Solving from Nature (PPSN)}} \emph{(\bibinfo{series}{LNCS},
  Vol.~\bibinfo{volume}{13398})}. \bibinfo{publisher}{Springer},
  \bibinfo{pages}{46--60}.
\newblock
\urldef\tempurl%
\url{https://doi.org/10.1007/978-3-031-14714-2\_4}
\showDOI{\tempurl}
\newblock
\shownote{Free version available at \url{https://arxiv.org/abs/2204.09483}}.


\bibitem[Lehre and Witt(2012)]%
        {LehreW12}
\bibfield{author}{\bibinfo{person}{Per~Kristian Lehre} {and}
  \bibinfo{person}{Carsten Witt}.} \bibinfo{year}{2012}\natexlab{}.
\newblock \showarticletitle{Black-Box Search by Unbiased Variation}.
\newblock \bibinfo{journal}{\emph{Algorithmica}}  \bibinfo{volume}{64}
  (\bibinfo{year}{2012}), \bibinfo{pages}{623--642}.
\newblock


\bibitem[L{\'{o}}pez{-}Ib{\'{a}}{\~{n}}ez et~al\mbox{.}(2021)]%
        {Reproducibility21TELO}
\bibfield{author}{\bibinfo{person}{Manuel L{\'{o}}pez{-}Ib{\'{a}}{\~{n}}ez},
  \bibinfo{person}{J{\"{u}}rgen Branke}, {and} \bibinfo{person}{Lu{\'{\i}}s
  Paquete}.} \bibinfo{year}{2021}\natexlab{}.
\newblock \showarticletitle{Reproducibility in Evolutionary Computation}.
\newblock \bibinfo{journal}{\emph{{ACM} Trans. Evol. Learn. Optim.}}
  \bibinfo{volume}{1}, \bibinfo{number}{4} (\bibinfo{year}{2021}),
  \bibinfo{pages}{14:1--14:21}.
\newblock
\urldef\tempurl%
\url{https://doi.org/10.1145/3466624}
\showDOI{\tempurl}


\bibitem[L{\'{o}}pez{-}Ib{\'{a}}{\~{n}}ez et~al\mbox{.}(2016)]%
        {irace}
\bibfield{author}{\bibinfo{person}{Manuel L{\'{o}}pez{-}Ib{\'{a}}{\~{n}}ez},
  \bibinfo{person}{J{\'{e}}r{\'{e}}mie Dubois{-}Lacoste},
  \bibinfo{person}{Leslie~P{\'{e}}rez C{\'{a}}ceres}, \bibinfo{person}{Mauro
  Birattari}, {and} \bibinfo{person}{Thomas St{\"{u}}tzle}.}
  \bibinfo{year}{2016}\natexlab{}.
\newblock \showarticletitle{The irace package: Iterated racing for automatic
  algorithm configuration}.
\newblock \bibinfo{journal}{\emph{Operations Research Perspectives}}
  \bibinfo{volume}{3} (\bibinfo{year}{2016}), \bibinfo{pages}{43--58}.
\newblock


\bibitem[Pushak and Hoos(2018)]%
        {PushakH18PPSN}
\bibfield{author}{\bibinfo{person}{Yasha Pushak} {and}
  \bibinfo{person}{Holger~H. Hoos}.} \bibinfo{year}{2018}\natexlab{}.
\newblock \showarticletitle{Algorithm Configuration Landscapes: - More Benign
  Than Expected?}. In \bibinfo{booktitle}{\emph{Proc. of Parallel Problem
  Solving from Nature}} \emph{(\bibinfo{series}{LNCS},
  Vol.~\bibinfo{volume}{11102})}. \bibinfo{publisher}{Springer},
  \bibinfo{pages}{271--283}.
\newblock
\urldef\tempurl%
\url{https://doi.org/10.1007/978-3-319-99259-4\_22}
\showDOI{\tempurl}


\bibitem[Pushak and Hoos(2022)]%
        {PushakH22telo}
\bibfield{author}{\bibinfo{person}{Yasha Pushak} {and}
  \bibinfo{person}{Holger~H. Hoos}.} \bibinfo{year}{2022}\natexlab{}.
\newblock \showarticletitle{AutoML Loss Landscapes}.
\newblock \bibinfo{journal}{\emph{{ACM} Trans. Evol. Learn. Optim.}}
  \bibinfo{volume}{2}, \bibinfo{number}{3} (\bibinfo{year}{2022}),
  \bibinfo{pages}{10:1--10:30}.
\newblock
\urldef\tempurl%
\url{https://doi.org/10.1145/3558774}
\showDOI{\tempurl}


\bibitem[Ros and Hansen(2008)]%
        {sep-cma-es}
\bibfield{author}{\bibinfo{person}{Raymond Ros} {and} \bibinfo{person}{Nikolaus
  Hansen}.} \bibinfo{year}{2008}\natexlab{}.
\newblock \showarticletitle{A simple modification in {CMA-ES} achieving linear
  time and space complexity}.
\newblock In \bibinfo{booktitle}{\emph{Parallel Problem Solving from Nature --
  {PPSN} {X}}}. Number 5199 in \bibinfo{series}{Lecture Notes in Computer
  Science}. \bibinfo{pages}{296--305}.
\newblock


\bibitem[Schulman et~al\mbox{.}(2017)]%
        {SchulmanWDRK17PPO}
\bibfield{author}{\bibinfo{person}{John Schulman}, \bibinfo{person}{Filip
  Wolski}, \bibinfo{person}{Prafulla Dhariwal}, \bibinfo{person}{Alec Radford},
  {and} \bibinfo{person}{Oleg Klimov}.} \bibinfo{year}{2017}\natexlab{}.
\newblock \showarticletitle{Proximal Policy Optimization Algorithms}.
\newblock \bibinfo{journal}{\emph{CoRR}}  \bibinfo{volume}{abs/1707.06347}
  (\bibinfo{year}{2017}).
\newblock
\urldef\tempurl%
\url{http://arxiv.org/abs/1707.06347}
\showURL{%
\tempurl}


\bibitem[Shala et~al\mbox{.}(2020)]%
        {shala-ppsn20}
\bibfield{author}{\bibinfo{person}{Gresa Shala}, \bibinfo{person}{André
  Biedenkapp}, \bibinfo{person}{Noor Awad}, \bibinfo{person}{Steven
  Adriaensen}, \bibinfo{person}{Marius Lindauer}, {and} \bibinfo{person}{Frank
  Hutter}.} \bibinfo{year}{2020}\natexlab{}.
\newblock \showarticletitle{Learning Step-Size Adaptation in {CMA-ES}}. In
  \bibinfo{booktitle}{\emph{Proc. of Parallel Problem Solving from Nature
  (PPSN'20)}} \emph{(\bibinfo{series}{LNCS}, Vol.~\bibinfo{volume}{12270})}.
  \bibinfo{publisher}{Springer}, \bibinfo{pages}{691--706}.
\newblock


\bibitem[Sharma et~al\mbox{.}(2019)]%
        {ManuelGECCO2019DDQN}
\bibfield{author}{\bibinfo{person}{Mudita Sharma}, \bibinfo{person}{Alexandros
  Komninos}, \bibinfo{person}{Manuel L{\'o}pez-Ib{\'a}{\~n}ez}, {and}
  \bibinfo{person}{Dimitar Kazakov}.} \bibinfo{year}{2019}\natexlab{}.
\newblock \showarticletitle{Deep Reinforcement Learning-Based Parameter Control
  in Differential Evolution}.
\newblock In \bibinfo{booktitle}{\emph{Proc. of Genetic and Evolutionary
  Computation Conference (GECCO'19)}}. \bibinfo{publisher}{ACM},
  \bibinfo{pages}{709--717}.
\newblock
\urldef\tempurl%
\url{https://doi.org/10.1145/3321707.3321813}
\showDOI{\tempurl}


\bibitem[Speck et~al\mbox{.}(2021)]%
        {BiedenkappHeuristicSelection}
\bibfield{author}{\bibinfo{person}{David Speck}, \bibinfo{person}{Andr{\'{e}}
  Biedenkapp}, \bibinfo{person}{Frank Hutter}, \bibinfo{person}{Robert
  Mattm{\"{u}}ller}, {and} \bibinfo{person}{Marius Lindauer}.}
  \bibinfo{year}{2021}\natexlab{}.
\newblock \showarticletitle{Learning Heuristic Selection with Dynamic Algorithm
  Configuration}. In \bibinfo{booktitle}{\emph{Proc. of International
  Conference on Automated Planning and Scheduling (ICAPS)}}.
  \bibinfo{publisher}{{AAAI} Press}, \bibinfo{pages}{597--605}.
\newblock
\urldef\tempurl%
\url{https://ojs.aaai.org/index.php/ICAPS/article/view/16008}
\showURL{%
\tempurl}


\bibitem[Tessari and Iacca(2022)]%
        {tessari2022reinforcement}
\bibfield{author}{\bibinfo{person}{Michele Tessari} {and}
  \bibinfo{person}{Giovanni Iacca}.} \bibinfo{year}{2022}\natexlab{}.
\newblock \showarticletitle{Reinforcement learning based adaptive
  metaheuristics}. In \bibinfo{booktitle}{\emph{Proceedings of the Genetic and
  Evolutionary Computation Conference Companion}}. \bibinfo{pages}{1854--1861}.
\newblock


\bibitem[Van~Hasselt et~al\mbox{.}(2016)]%
        {van2016deep}
\bibfield{author}{\bibinfo{person}{Hado Van~Hasselt}, \bibinfo{person}{Arthur
  Guez}, {and} \bibinfo{person}{David Silver}.}
  \bibinfo{year}{2016}\natexlab{}.
\newblock \showarticletitle{Deep reinforcement learning with double
  q-learning}. In \bibinfo{booktitle}{\emph{Proceedings of the AAAI conference
  on artificial intelligence}}, Vol.~\bibinfo{volume}{30}.
\newblock


\bibitem[Xue et~al\mbox{.}(2022)]%
        {xue2022multi}
\bibfield{author}{\bibinfo{person}{Ke Xue}, \bibinfo{person}{Jiacheng Xu},
  \bibinfo{person}{Lei Yuan}, \bibinfo{person}{Miqing Li},
  \bibinfo{person}{Chao Qian}, \bibinfo{person}{Zongzhang Zhang}, {and}
  \bibinfo{person}{Yang Yu}.} \bibinfo{year}{2022}\natexlab{}.
\newblock \showarticletitle{Multi-agent Dynamic Algorithm Configuration}. In
  \bibinfo{booktitle}{\emph{Advances in Neural Information Processing Systems
  35 (NeurIPS'22)}}. \bibinfo{address}{New Orleans, LA}.
\newblock


\bibitem[Yi et~al\mbox{.}(2023)]%
        {yi2023automated}
\bibfield{author}{\bibinfo{person}{Wenjie Yi}, \bibinfo{person}{Rong Qu}, {and}
  \bibinfo{person}{Licheng Jiao}.} \bibinfo{year}{2023}\natexlab{}.
\newblock \showarticletitle{Automated algorithm design using proximal policy
  optimisation with identified features}.
\newblock \bibinfo{journal}{\emph{Expert Systems with Applications}}
  \bibinfo{volume}{216} (\bibinfo{year}{2023}), \bibinfo{pages}{119461}.
\newblock


\bibitem[Yi et~al\mbox{.}(2022)]%
        {yi2022automated}
\bibfield{author}{\bibinfo{person}{Wenjie Yi}, \bibinfo{person}{Rong Qu},
  \bibinfo{person}{Licheng Jiao}, {and} \bibinfo{person}{Ben Niu}.}
  \bibinfo{year}{2022}\natexlab{}.
\newblock \showarticletitle{Automated Design of Metaheuristics Using
  Reinforcement Learning within a Novel General Search Framework}.
\newblock \bibinfo{journal}{\emph{IEEE Transactions on Evolutionary
  Computation}} (\bibinfo{year}{2022}).
\newblock


\bibitem[Zhang et~al\mbox{.}(2022)]%
        {zhang2022deep}
\bibfield{author}{\bibinfo{person}{Yuchang Zhang}, \bibinfo{person}{Ruibin
  Bai}, \bibinfo{person}{Rong Qu}, \bibinfo{person}{Chaofan Tu}, {and}
  \bibinfo{person}{Jiahuan Jin}.} \bibinfo{year}{2022}\natexlab{}.
\newblock \showarticletitle{A deep reinforcement learning based hyper-heuristic
  for combinatorial optimisation with uncertainties}.
\newblock \bibinfo{journal}{\emph{European Journal of Operational Research}}
  \bibinfo{volume}{300}, \bibinfo{number}{2} (\bibinfo{year}{2022}),
  \bibinfo{pages}{418--427}.
\newblock


\end{thebibliography}

\end{document}